\documentclass{article}

\usepackage{PRIMEarxiv}

\usepackage[utf8]{inputenc} % allow utf-8 input
\usepackage[T1]{fontenc}    % use 8-bit T1 fonts
\usepackage{hyperref}       % hyperlinks
\usepackage{url}            % simple URL typesetting
\usepackage{booktabs}       % professional-quality tables
\usepackage{amsfonts}       % blackboard math symbols
\usepackage{nicefrac}       % compact symbols for 1/2, etc.
\usepackage{microtype}      % microtypography
\usepackage{lipsum}
\usepackage{fancyhdr}       % header
\usepackage{graphicx}       % graphics
\usepackage{enumitem}
\usepackage{caption}
\usepackage{wrapfig}
\usepackage{pdfpages}
\usepackage[affil-it]{authblk}  % Required for author/affiliation formatting
\usepackage{hyperref} % Optional for clickable emails

\graphicspath{{media/}}     % organize your images and other figures under media/ folder
\title{Automatic Personalized Impression Generation for PET Reports Using Large Language Models}
\author[1,2]{Xin Tie}
\author[1]{Muheon Shin}
\author[1,2]{Ali Pirasteh}
\author[1]{Nevein Ibrahim}
\author[1]{Zachary Huemann}
\author[3,4]{Sharon M. Castellino}
\author[5,6]{\\Kara M. Kelly}
\author[1,2]{John Garrett}
\author[7,8]{Junjie Hu}
\author[1,9]{Steve Y. Cho}
\author[$\enspace$1]{Tyler J. Bradshaw\thanks{Corresponding author: \href{mailto:tbradshaw@wisc.edu}{tbradshaw@wisc.edu}}}

\affil[1]{\hspace{0.8em}Department of Radiology, University of Wisconsin, Madison, WI, USA}
\affil[2]{\hspace{0.8em}Department of Medical Physics, University of Wisconsin, Madison, WI, USA}
\affil[3]{\hspace{0.8em}Department of Pediatrics, Emory University School of Medicine, Atlanta, GA, USA}
\affil[4]{\hspace{0.8em}Aflac Cancer and Blood Disorders Center, Children's Healthcare of Atlanta, Atlanta, GA, USA}
\affil[5]{\hspace{0.8em}Department of Pediatric Oncology, Roswell Park Comprehensive Cancer Center, Buffalo, NY, USA}
\affil[6]{\hspace{0.8em}Department of Pediatrics, University at Buffalo Jacobs School of Medicine and Biomedical Sciences, Buffalo, NY, USA}
\affil[7]{\hspace{0.8em}Department of Biostatistics and Medical Informatics, University of Wisconsin, Madison, WI, USA}
\affil[8]{\hspace{0.8em}Department of Computer Science, University of Wisconsin, Madison, WI, USA}
\affil[9]{\hspace{0.8em}University of Wisconsin Carbone Comprehensive Cancer Center, Madison, WI, USA}

\begin{document}
\maketitle
\thispagestyle{plain}  % Forces the first page to have a page number
\begin{abstract}
\textbf{Purpose}: To determine if fine-tuned large language models (LLMs) can generate accurate, personalized impressions for whole-body PET reports. \\
\textbf{Materials and Methods}: Twelve language models were trained on a corpus of PET reports using the teacher-forcing algorithm, with the report findings as input and the clinical impressions as reference. An extra input token encodes the reading physician’s identity, allowing models to learn physician-specific reporting styles. Our corpus comprised 37,370 retrospective PET reports collected from our institution between 2010 and 2022. To identify the best LLM, 30 evaluation metrics were benchmarked against quality scores from two nuclear medicine (NM) physicians, with the most aligned metrics selecting the model for expert evaluation. In a subset of data, model-generated impressions and original clinical impressions were assessed by three NM physicians according to 6 quality dimensions (3-point scale) and an overall utility score (5-point scale). Each physician reviewed 12 of their own reports and 12 reports from other physicians. Bootstrap resampling was used for statistical analysis. \\
\textbf{Results}: Of all evaluation metrics, domain-adapted BARTScore and PEGASUSScore showed the highest Spearman’s $\rho$ correlations ($\rho$=0.568 and 0.563) with physician preferences. Based on these metrics, the fine-tuned PEGASUS model was selected as the top LLM. When physicians reviewed PEGASUS-generated impressions in their own style, 89$\%$ were considered clinically acceptable, with a mean utility score of 4.08 out of 5. Physicians rated these personalized impressions as comparable in overall utility to the impressions dictated by other physicians (4.03, $P$=0.41).\\
\textbf{Conclusion}: Personalized impressions generated by PEGASUS were clinically useful, highlighting its potential to expedite PET reporting.
\end{abstract}

% keywords can be removed

\section{Introduction}
The radiology report serves as the official interpretation of a radiological examination and is essential for communicating relevant clinical findings amongst reading physicians, the healthcare team, and patients. Compared to other imaging modalities, reports for whole-body PET exams (e.g., skull base to thigh or skull vertex to feet) are notable for their length and complexity (1). In a typical PET report, the findings section details numerous observations about the study and the impression section summarizes the key findings. Given that referring physicians primarily rely on the impression section for clinical decision-making and management (2), it is paramount to ensure its accuracy and completeness. However, creating PET impressions that encapsulate all important findings can be time-consuming and error-prone (3). Large language models (LLMs) have the potential to accelerate this process by automatically drafting impressions based on the findings.

While LLMs have been used previously to summarize radiology findings (3–8), impression generation for whole-body PET reports has received comparatively little attention. Unlike general chest x-ray reports that comprise 40-70 words (9) in the findings section, whole-body PET reports often have 250-500 words in the findings section and contain observations across multiple anatomical regions with cross comparison to available anatomic imaging modalities (e.g., CT and MRI) and clinical correlation. This complexity heightens the risk of omissions in the generated impressions. Furthermore, the length of PET impressions can accentuate the unique reporting styles of individual reading physicians, underscoring the need for personalized impression generation. Consequently, adapting LLMs for PET report summarization presents distinct challenges.

Evaluating the performances of LLMs in the task of impression generation is also challenging, given that a single case can have various acceptable impressions. While expert evaluation stands as the gold standard, it is impractical for physicians to exhaustively review outputs of all LLMs to determine the leading model. In recent years, several evaluation metrics designed for general text summarization have been adapted to evaluate summaries of medical documents (10-12). However, it remains unclear how well these metrics could assess the quality of PET impressions and which of them align most closely with physician judgments.

This study aimed to determine whether the LLMs fine-tuned on a large corpus of PET clinical reports can accurately summarize PET findings and generate impressions suitable for clinical use. We benchmarked 30 evaluation metrics against physician preferences and then selected the top-performing LLM. To assess its clinical utility, we conducted an expert reader study, identifying common mistakes made by the LLM. We also investigated the importance of personalizing impressions for reading physicians. Lastly, we evaluated the model’s reasoning capability within the nuclear medicine (NM) domain and performed external testing of our fine-tuned LLM.

\section{Materials and Methods}
\label{sec:headings}
\subsection{Dataset Collection}
Under a protocol approved by the institutional review board and with a waiver of informed consent, we collected 37,370 retrospective PET reports, dictated by 65 physicians, from our institution between January 2010 and January 2023. Appendix S1 presents the statistics of PET reports in our dataset. All reports were anonymized using NLM-Scrubber (13). Of 37,370 PET reports, 4000 were randomly selected for internal testing, 2000 were used for validation, and the remaining 31,370 reports were used for training. For external testing, we retrieved 100 whole-body PET/CT reports, dictated by 62 physicians, from Children’s Oncology Group AHOD1331 Hodgkin lymphoma clinical trial (ClinicalTrials.gov number, NCT02166463) (14). There is no overlap between physicians in the internal and external sets.

\subsection{Report Preprocessing}
\begin{figure}[h!]
\centering
\includegraphics[width=0.99\textwidth]{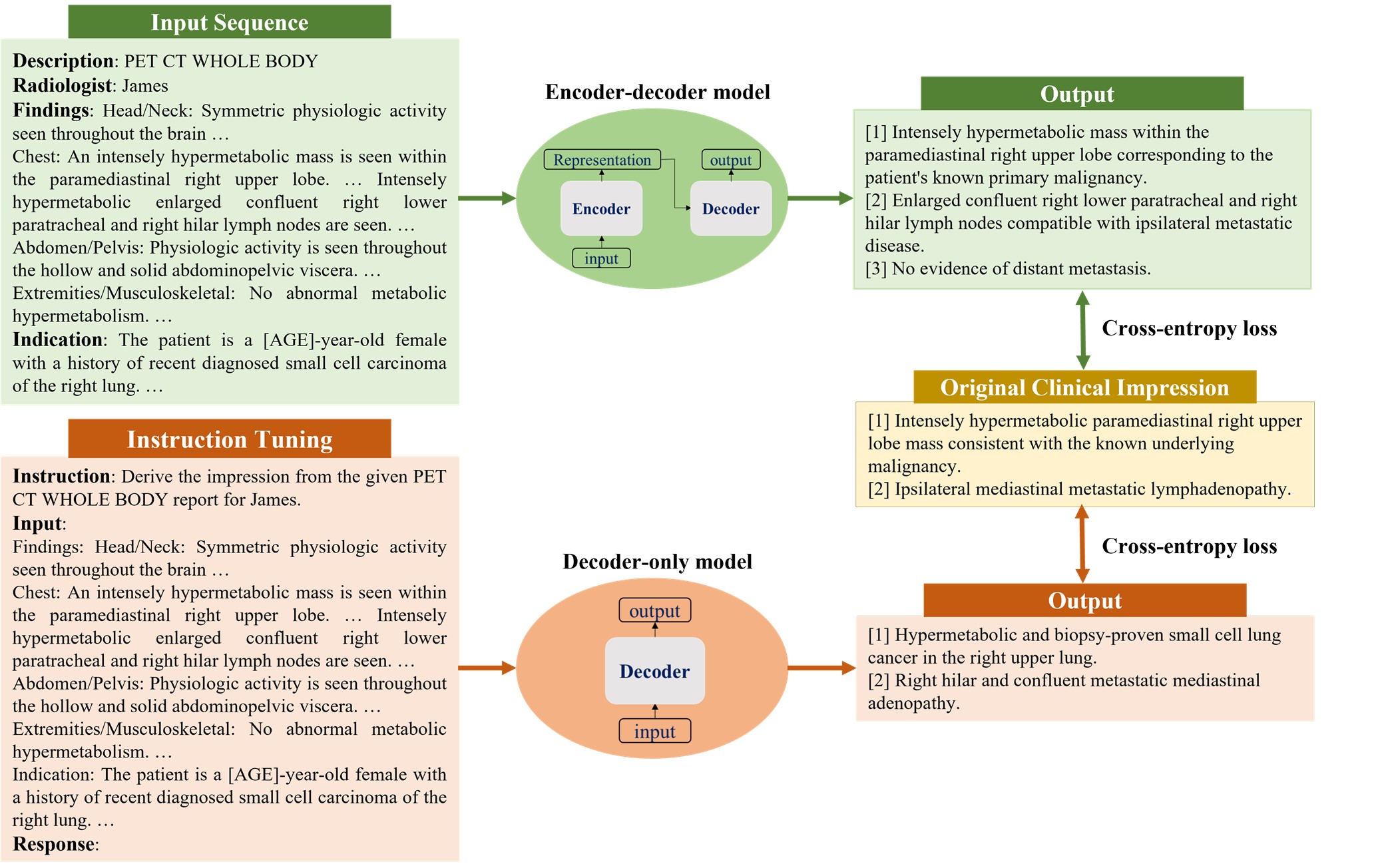}
  \caption{\small{Formatting of reports for input to encoder-decoder and decoder-only models. For encoder-decoder models, the first two lines describe the examination category and encode the reading physician’s identity, respectively. “Findings” contains the clinical findings from the PET report, and “Indication” includes the patient’s medical history and the reason for the examination. For decoder-only models, each case follows a specific format for the instruction: “Derive the impression from the given [description] for [physician]”. “Input” accommodates the concatenation of clinical findings and indications. The output always starts with the prefix “Response:”. Both model architectures utilize the cross-entropy loss to compute the difference between original clinical impressions and model-generated impressions.}} 
  \vspace{-10pt}
  \label{fig:fig1}
\end{figure}

In this work, we investigated both encoder-decoder and decoder-only language models. Considering their different architectures, we customized input templates as illustrated in Figure \ref{fig:fig1}. For encoder-decoder models, the first lines describe the categories of PET scans, while the second lines encode each reading physician’s identity using an identifier token (details in Appendix S2). The “Findings” section contains the clinical findings from the PET reports, whereas the “Indications” section encompasses relevant background information, including the patient’s medical history and the reason for the examination. For decoder-only models, we employed the instruction-tuning method (15) and adapted the prompt from (16). Each case starts with the instruction: “Derive the impression from the given [description] report for [physician].” The PET findings and background information are concatenated to form the “Input” section. The original clinical impressions are used as the reference for model training and evaluation.

\subsection{Models for PET Report Summarization}
We formulated our work as an abstractive summarization task since physicians typically interpret findings in the impression section, rather than merely reusing sentences from the findings section. We fine-tuned 8 encoder-decoder models and 4 decoder-only models, covering a broad range of language models for sequence generation. The encoder-decoder models comprised state-of-the-art (SOTA) transformer-based models, namely BART (17), PEGASUS (18), T5 (19) and FLAN-T5 (20). To investigate if the medical-domain adaptation could benefit our task, we fine-tuned 2 biomedical LLMs, BioBART (21) and Clinical-T5 (22). Additionally, we included 2 baseline models, pointer-generator network (PGN) (3) and BERT2BERT (23). The decoder-only models encompassed GPT2 (24) and OPT (25) as well as the recently released LLaMA (26) and Alpaca (16). All models were trained using the standard teacher-forcing algorithm. LLaMA and Alpaca were fine-tuned with low-rank adaptation (LoRA) (27) to reduce memory usage and accelerate training, while the other models were subjected to full fine-tuning. We adopted the beam search decoding algorithm to generate impressions. More comprehensive information about these models, including their training and inference settings, can be found in Appendix S3. 

Our models are made available on Hugging Face: \url{https://huggingface.co/xtie/PEGASUS-PET-impression}. The code can be found in the open-source project: \url{https://github.com/xtie97/PET-Report-Summarization}.

\subsection{Benchmarking Evaluation Metrics}

To identify the evaluation metrics most correlated with physician preferences, we presented impressions generated by 4 different models (PGN, BERT2BERT, BART, PEGASUS) to two NM physicians. These models represented a wide performance spectrum. One physician (M.S.) reviewed 200 randomly sampled reports in the test set, then scored the quality of model-generated impressions on a 5-point Likert scale. The definitions of each level are provided in Appendix S4. To assess inter-observer variability, a second physician (S.Y.C.) independently scored 20 of the cases based on the same criterion. 

Table \ref{table:table1} categorizes the evaluation metrics (detailed introductions in Appendix S4) included in this study. To address the domain gap between general-domain articles and PET reports, we fine-tuned BARTScore on our PET reports using the method described in (28) and named it BARTScore$+$PET. Following the same approach, we developed PEGASUSScore$+$PET and T5Score$+$PET. These three evaluators are made available at \url{https://huggingface.co/xtie/BARTScore-PET}. The Spearman’s $\rho$ correlation quantified how well evaluation metrics correlated with the physicians’ judgments. Metrics with the highest correlations were used to determine the top-performing model.

\addtocounter{figure}{-1} % Decrement the figure counter
\begin{figure}[h!]
\centering
\captionsetup{name=Table}
\caption{\small{All evaluation metrics included in this study and their respective categories.}}
\includegraphics[width=0.98\textwidth]{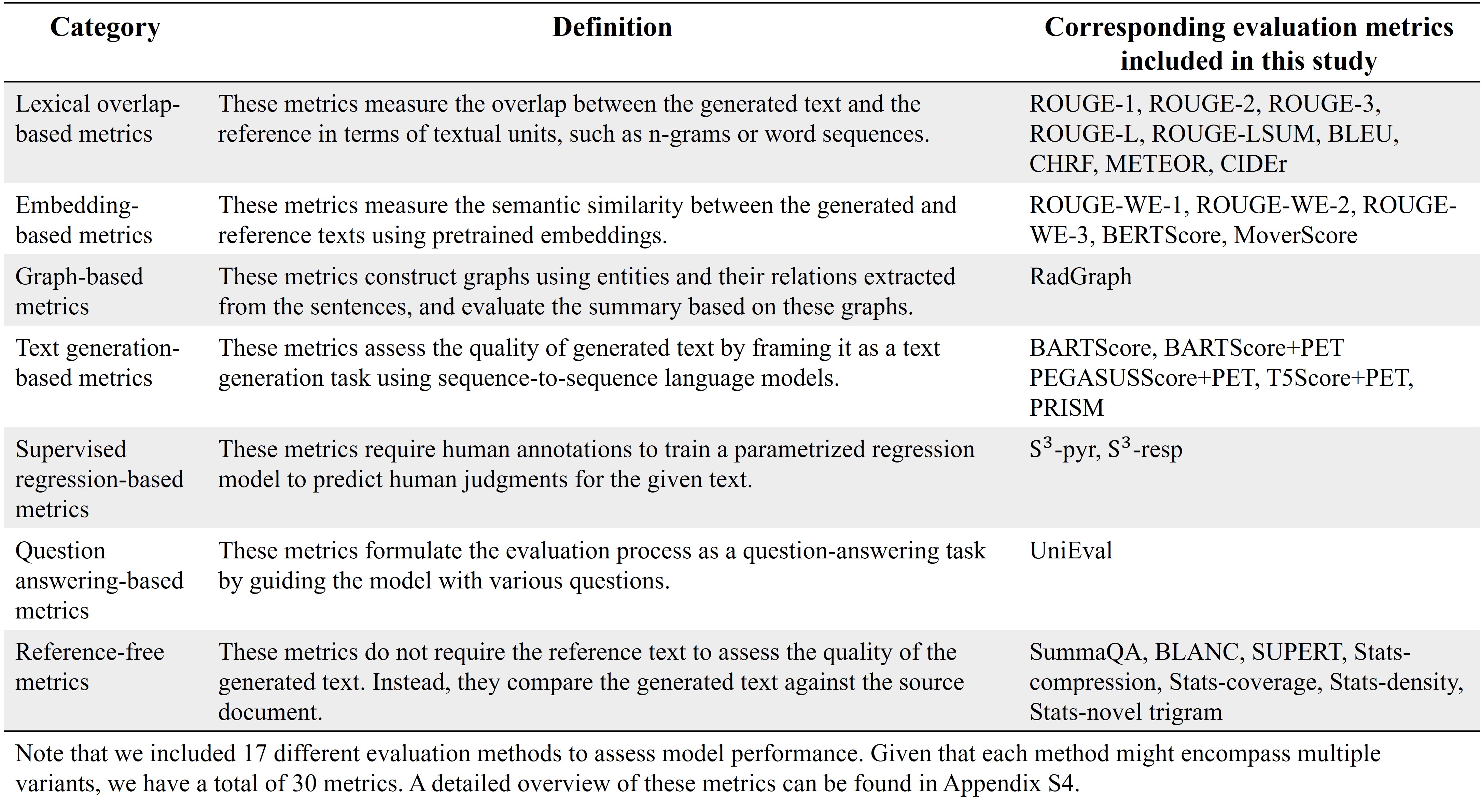}
  \vspace{-10pt}
\label{table:table1}
\end{figure}

\subsection{Expert Evaluation}
\begin{figure}[h!]
\centering
\captionsetup{name=Table}
\caption{\small{Definitions of six quality dimensions and an overall utility score used in our expert evaluation, along with their corresponding Likert systems.}}
\includegraphics[width=0.99\textwidth]{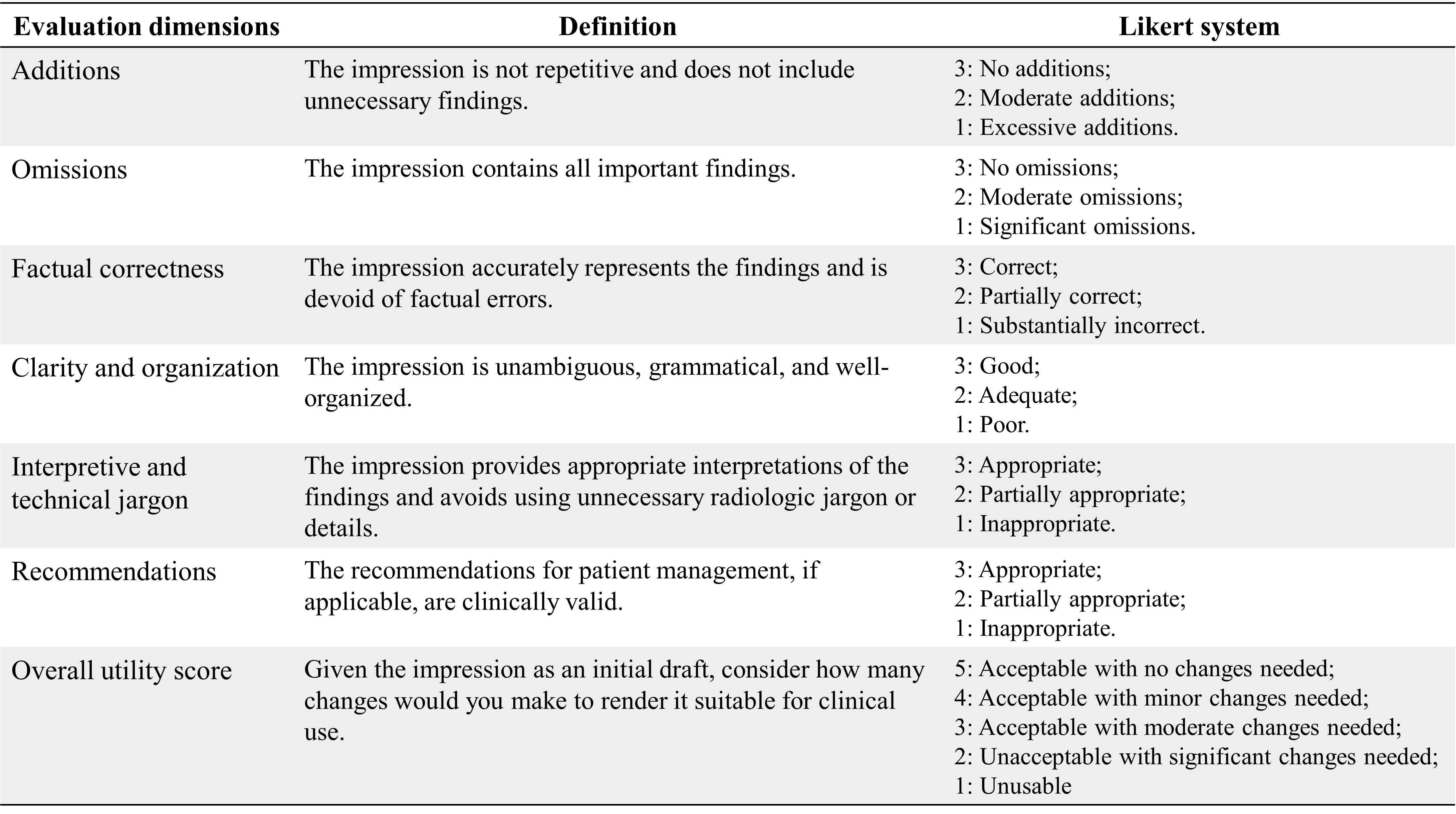}
  \vspace{-5pt}
\label{table:table2}
\end{figure}

To examine the clinical utility of our best LLM, we conducted a reader study involving three physicians: two board-certified in NM (N.I., S.Y.C.) and one board-certified in NM and radiology (A.P.). Blinded to the original interpreting physicians, each reader independently reviewed a total of 24 whole-body PET reports along with model-generated impressions. Of these, twelve cases were originally dictated by themselves, and the rest were dictated by other physicians. The LLM impressions were always generated in the style of the interpreting physician by using their specific identifier token. The scoring system included 6 quality dimensions (3-point scale) and an overall utility score (5-point scale). Their definitions are described in Table \ref{table:table2}. The application we designed for physician review of test cases can be accessed at \url{https://github.com/xtie97/PET-Report-Expert-Evaluation}.

\subsection{Additional Analysis}
To further evaluate the capability of our fine-tuned LLMs, we conducted three additional experiments. Implementation details are provided in Appendix S5:
\begin{itemize}[leftmargin=13pt]
    \item[1.] \textbf{Deauville score (DS) prediction}: To test the reasoning capability of our models within the NM domain, we classified PET lymphoma reports into DS 1-5 based on the exam-level DSs (29) extracted from model-generated impressions. The original clinical impressions served as the reference for the DSs. The evaluation metrics included the 5-class accuracy and the linearly weighted Cohen’s $\kappa$ index. For context, a prior study (29) showed that a human expert predicted DSs with 66$\%$ accuracy and a Cohen’s $\kappa$ of 0.79 when the redacted PET reports and maximum intensity projection images were given.

    \item[2.] \textbf{Encoding physician-specific styles}: We compared the impressions generated in the styles of two physicians (Physician 1 and Physician 2) who had distinct reporting styles. Physician 1’s impressions tended to be more detailed, whereas Physician 2’s impressions were more concise. 

    \item[3.] \textbf{External testing}: We generated the impressions for all external cases in the styles of three primary physicians (Physician 1, Physician 2, and Physician 3) from our internal dataset and compared these impressions with clinical impressions originally dictated by external physicians.  
\end{itemize}

\subsection{Statistical Analysis}
Using bootstrap resampling (30), the 95$\%$ confidence intervals (CI) for our results were derived from 10,000 repetitive trials. The difference between two data groups was statistically significant at 0.05 only when one group exceeded the other in 95$\%$ of trials.

\section{Results}
\subsection{Benchmarking evaluation metrics}
Figure \ref{fig:fig2} shows the Spearman's $\rho$ correlation between evaluation metrics and quality scores assigned by the first physician (M.S.). BARTScore+PET and PEGASUSScore+PET exhibited the highest correlations with physician judgment ($\rho$=0.568 and 0.563, $P$=0.30). Therefore, both metrics were employed to determine the top-performing model for expert evaluation. However, their correlation values were still below the degree of inter-reader correlation ($\rho$=0.654). Similar results were observed in the correlation between evaluation metrics and the second physician’s scores (Appendix S6). Without adaption to PET reports, the original BARTScore showed lower correlation ($\rho$=0.474, $P<$0.001) compared to BARTScore+PET, but still outperformed traditional evaluation metrics like Recall-Oriented Understudy for Gisting Evaluation-L (ROUGE-L, $\rho$=0.398, $P<$0.001) (31). 

The metrics commonly used in radiology report summarization, including ROUGE (31), BERTScore (32) and RadGraph (10), did not demonstrate strong correlation with physician preferences. Additionally, most reference-free metrics, although effective in general text summarization, showed considerably lower correlation compared to reference-dependent metrics. 
 
\subsection{Model Performance}
Figure \ref{fig:fig3} illustrates the relative performance of 12 language models assessed using all evaluation metrics considered in this study. For better visualization, metric values have been normalized to [0, 1], with the original values available in Appendix S7. The SOTA encoder-decoder models, including PEGASUS, BART, and T5, demonstrated similar performance across most evaluation metrics. Since BARTScore+PET and PEGASUSScore+PET identified PEGASUS as the top-performing model, we selected it for further expert evaluation. 

\addtocounter{figure}{-1} % Decrement the figure counter
\begin{wrapfigure}{r}{0.47\textwidth}
\vspace{-8pt}
\centering 
\includegraphics[width=1.0\linewidth]{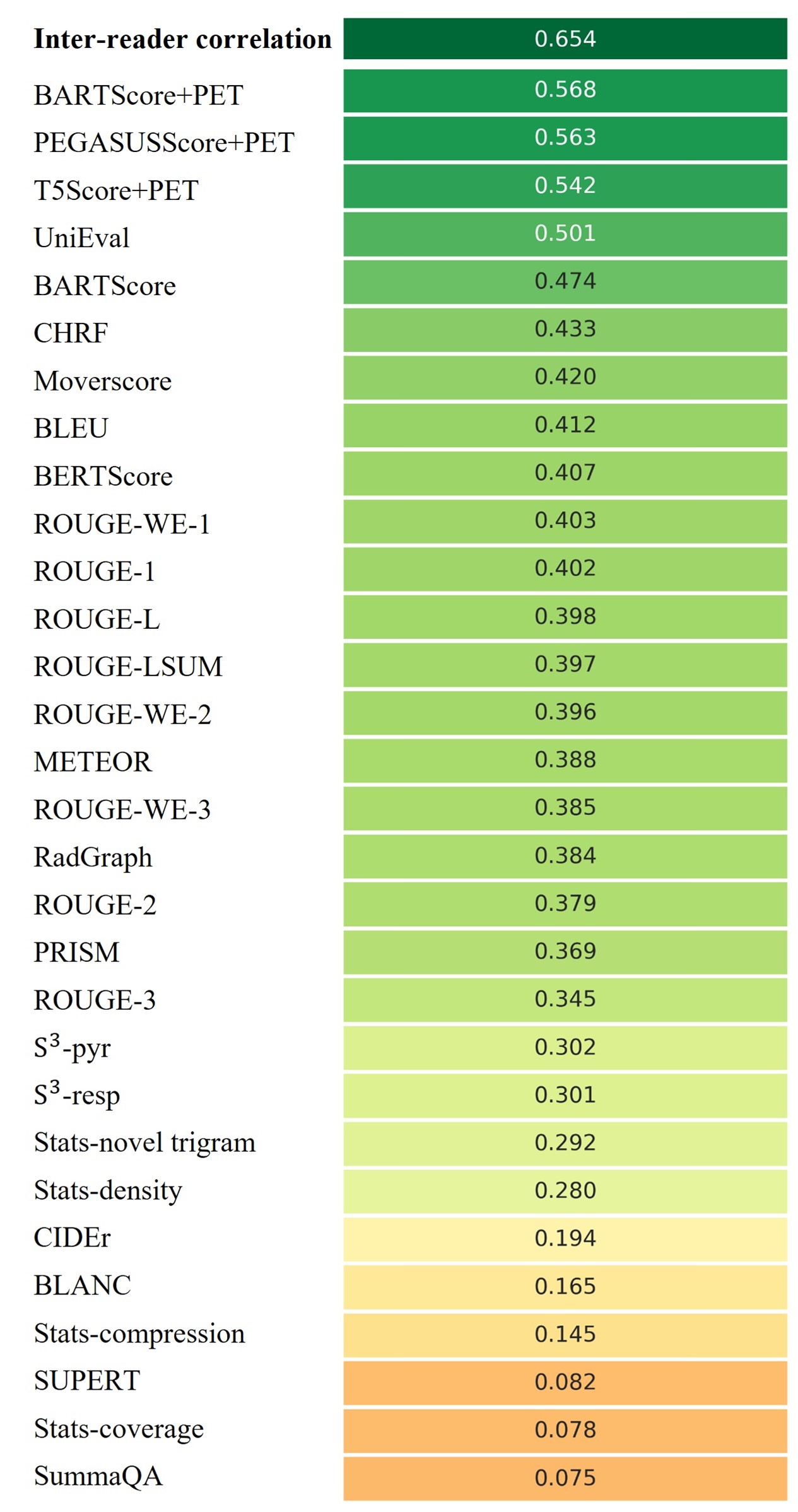} 
\caption{\small{Spearman’s $\rho$ correlations between different evaluation metrics and quality scores assigned by the first physician. The top row quantifies the inter-reader correlation. Notably, domain-adapted BARTScore (BARTScore+PET) and PEGASUSScore (PEGASUSScore+PET) demonstrate the highest correlations with physician preferences. }} 
\vspace{-25pt}
\label{fig:fig2}
\end{wrapfigure}

After being fine-tuned on our PET reports, the medical knowledge enriched models, BioBART (BARTScore+PET: -1.46; ROUGE-L: 38.9) and Clinical-T5 (BARTScore+PET: -1.54; ROUGE-L: 39.4), did not show superior performance compared to their base models, BART (BARTScore+PET: -1.46; ROUGE-L: 38.6) and T5 (BARTScore+PET: -1.52; ROUGE-L: 40.3). Additionally, the four decoder-only models included in this study showed significantly lower performance ($P<$0.001) compared to the top-tier encoder-decoder LLMs. Interestingly, LLaMA-LoRA (BARTScore+PET: -2.26; ROUGE-L: 27.2) and Alpaca-LoRA (BARTScore+PET: -2.24; ROUGE-L: 28.0), which have been pretrained on one trillion tokens, did not surpass the performance of GPT2 (BARTScore+PET: -2.04, ROUGE-L: 28.7) and OPT (BARTScore+PET: -2.07, ROUGE-L: 28.3).

\subsection{Expert Evaluation}
The distributions of overall utility scores and 6 specific quality scores are illustrated in Figure \ref{fig:fig4}. In total, 83$\%$ (60/72) of the PEGASUS-generated impressions were scored as clinically acceptable (scores 3-5), with 60$\%$ (43/72) scoring 4 or higher, and 28$\%$ (20/72) receiving a score of 5. When the physicians reviewed their own reports, 89$\%$ (32/36) of the PEGASUS-generated impressions were clinically acceptable, with a mean utility score of 4.08 (95$\%$ CI, 3.72, 4.42). This score was significantly ($P<$0.001) lower than the mean utility score (4.75, 95$\%$ CI, 4.58, 4.89) of the clinical impressions originally dictated by themselves. The discrepancy was primarily attributable to 3 quality dimensions: “factual correctness” (Clinical vs. PEGASUS: 2.97 vs. 2.58, $P$=0.001), “interpretive and technical jargon” (2.94 vs. 2.78, $P$=0.034) and “recommendations” (3.00 vs. 2.69, $P$=0.001). 

When the physicians evaluated clinical impressions dictated by other physicians, the mean utility score (4.03, 95$\%$ CI, 3.69, 4.33) was significantly lower than scores they assigned to their own impressions ($P<$0.001), suggesting a strong preference for their individual reporting style. The primary quality dimensions contributing to such difference included “additions” (Physician’s own impressions vs. Other physicians’ impressions: 2.94 vs. 2.75, $P$=0.039) and “clarity and organization” (2.92 vs. 2.50, $P<$0.001). On average, the physicians considered the overall utility of PEGASUS-generated impressions in their own style to be comparable to the clinical impressions dictated by other physicians (mean utility score: 4.08 vs. 4.03, $P$=0.41).

Figure \ref{fig:fig5} presents four PEGASUS-generated impressions (findings and background information in Appendix S8) with overall utility scores ranging from 2 to 5. For each case, PEGASUS successfully identified the salient findings, offered interpretations, and provided recommendations. However, the model showed susceptibility to factual incorrectness, including misinterpretation of findings and inconsistent statements in the impressions, as evidenced in case 4. Additionally, the model could give overly definite diagnoses, as observed in case 3. 

\subsection{Deauville Score Prediction}
\begin{figure}[h!]
\centering
\captionsetup{name=Table}
\caption{\small{Performance of 12 language models on Deauville score prediction}} 
\includegraphics[width=0.60\textwidth]{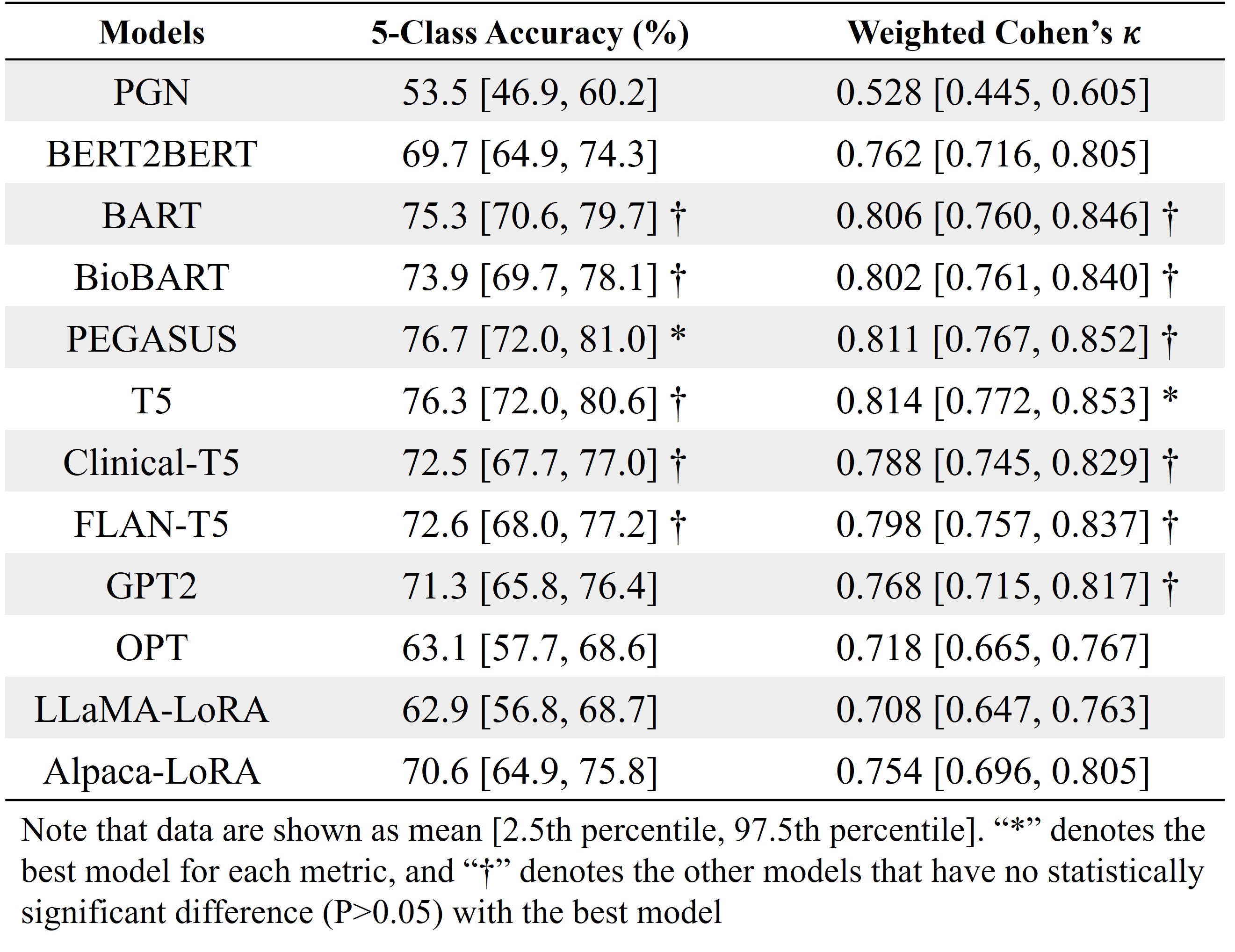}
\label{table:table3}
\end{figure}

Of the 4,000 test cases, 405 PET lymphoma reports contained DSs in the impression sections. Table \ref{table:table3} presents the DS classification results for all evaluated models. PEGASUS achieved the highest 5-class accuracy (76.7$\%$, 95$\%$ CI, 72.0$\%$, 81.0$\%$), while PGN was least effective in deriving DSs. All SOTA encoder-decoder models attained an accuracy exceeding 70$\%$. Among decoder-only models, GPT2 demonstrated the best performance, with an accuracy of 71.3$\%$ (95$\%$ CI, 65.8$\%$, 76.4$\%$).

\addtocounter{figure}{-1} % Decrement the figure counter

\begin{figure}[h!]
\vspace{-2pt}
\centering
\includegraphics[width=0.99\textwidth]{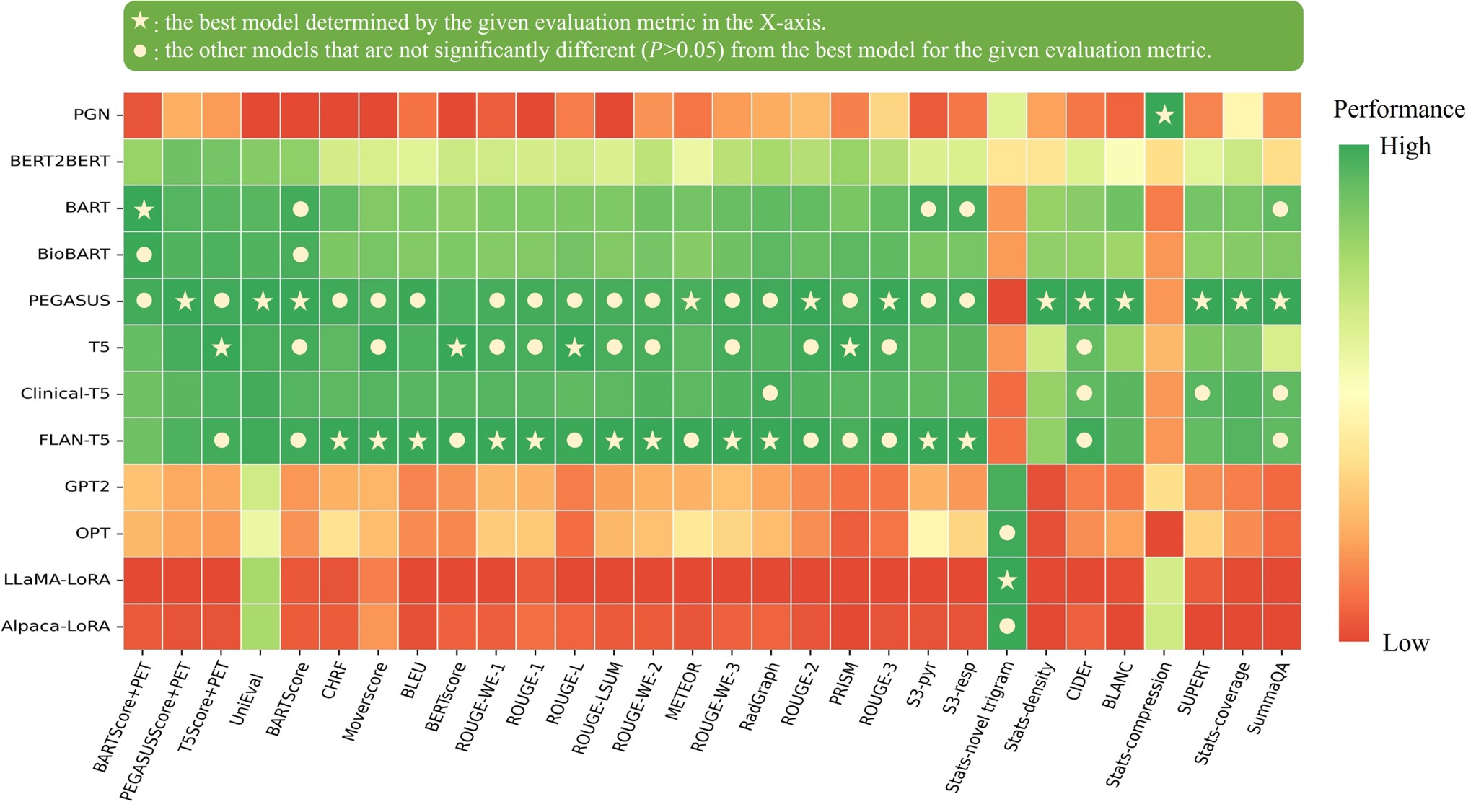}
  \caption{\small{Performance of 12 language models evaluated by the metrics included in this study. The X-axis displays the metrics arranged in descending order of correlation with physician preferences, with higher correlations on the left and lower correlations on the right. For each evaluation metric, values underwent min-max normalization to allow comparison within a single plot. The actual metric values are referenced in Appendix S7. The star denotes the best model for each metric, and the circle denotes the other models that do not have statistically significant difference ($P>$0.05) with the best model.}} 
    \vspace{-15pt}
  \label{fig:fig3}
\end{figure}

\subsection{Encoding Physician-specific Styles}

Figure \ref{fig:fig6} shows the PEGASUS-generated impressions given unique identifier tokens associated with two physicians, Physician 1 and Physician 2. Altering a single token in the input could lead to a drastic change in the output impressions. For each case, both impressions managed to capture the salient findings and delivered similar diagnoses, however, their length, level of detail and phrasing generally reflected the respective physician’s style. This reveals the model’s ability to tailor the impressions to individual physicians. The associated findings and background information are presented in Appendix S9. 

\begin{figure}[h!]
  \vspace{-10pt}
\centering
\includegraphics[width=0.99\textwidth]{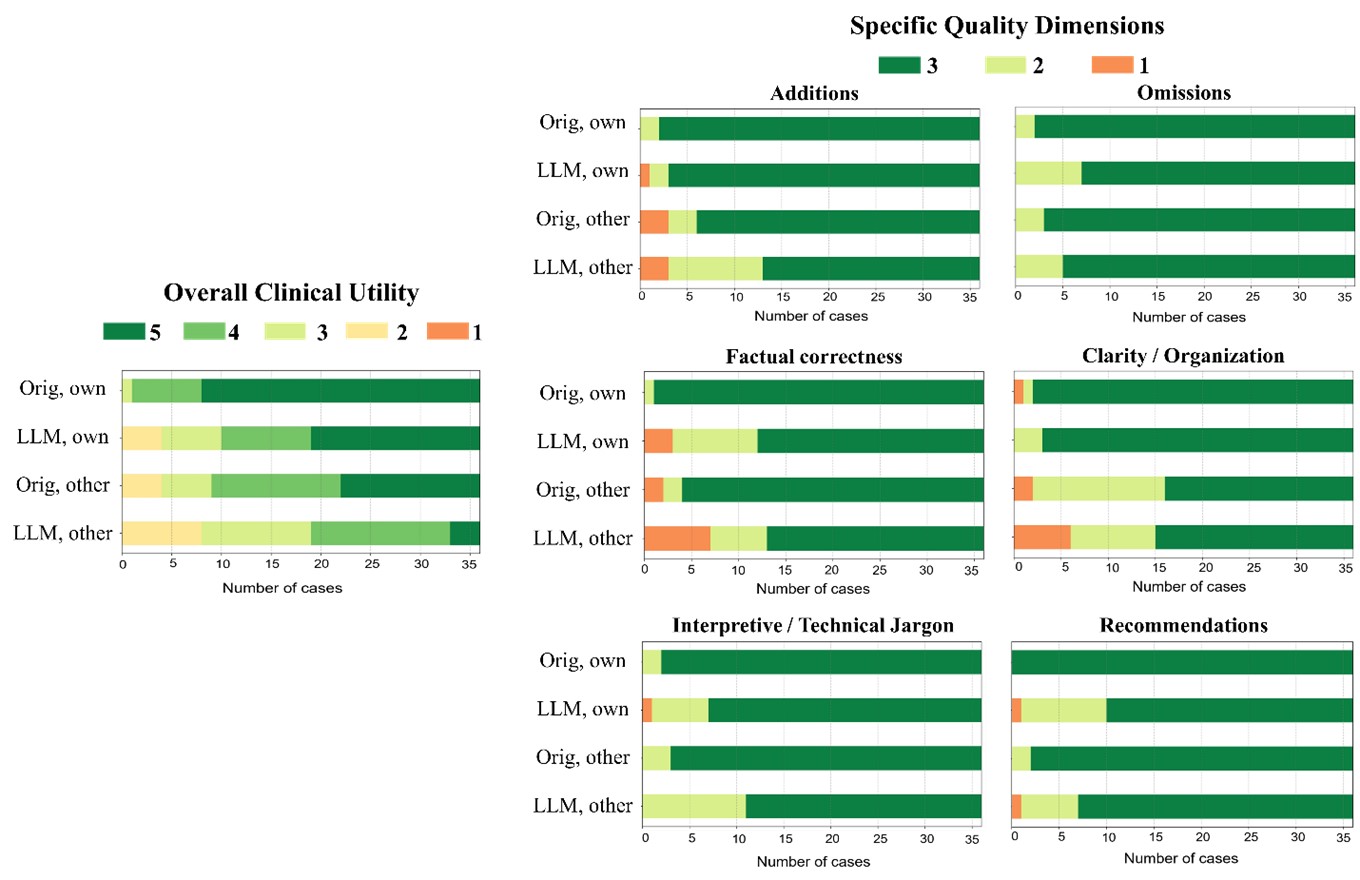}
  \caption{\small{Expert evaluation consisting of an overall utility score and 6 specific quality dimensions. For the physician’s own reports, 89$\%$ (32/36) of the PEGASUS-generated impressions were deemed clinically acceptable. The primary reasons for the discrepancy between original clinical impressions and PEGASUS-generated impressions are factual inaccuracies, inappropriate interpretations, and unsuitable recommendations. “Orig, own”: original clinical impressions from the physician’s own reports; “LLM, own”: PEGASUS-generated impressions for the physician’s own reports; “Orig, other”: original clinical impressions from other physicians’ reports; “LLM, other”: PEGASUS-generated impressions for other physicians’ reports.}} 
  \label{fig:fig4}
  \vspace{-4pt}
\end{figure}

\begin{figure}[h!]
\centering
  \vspace{-10pt}
\includegraphics[width=0.98\textwidth]{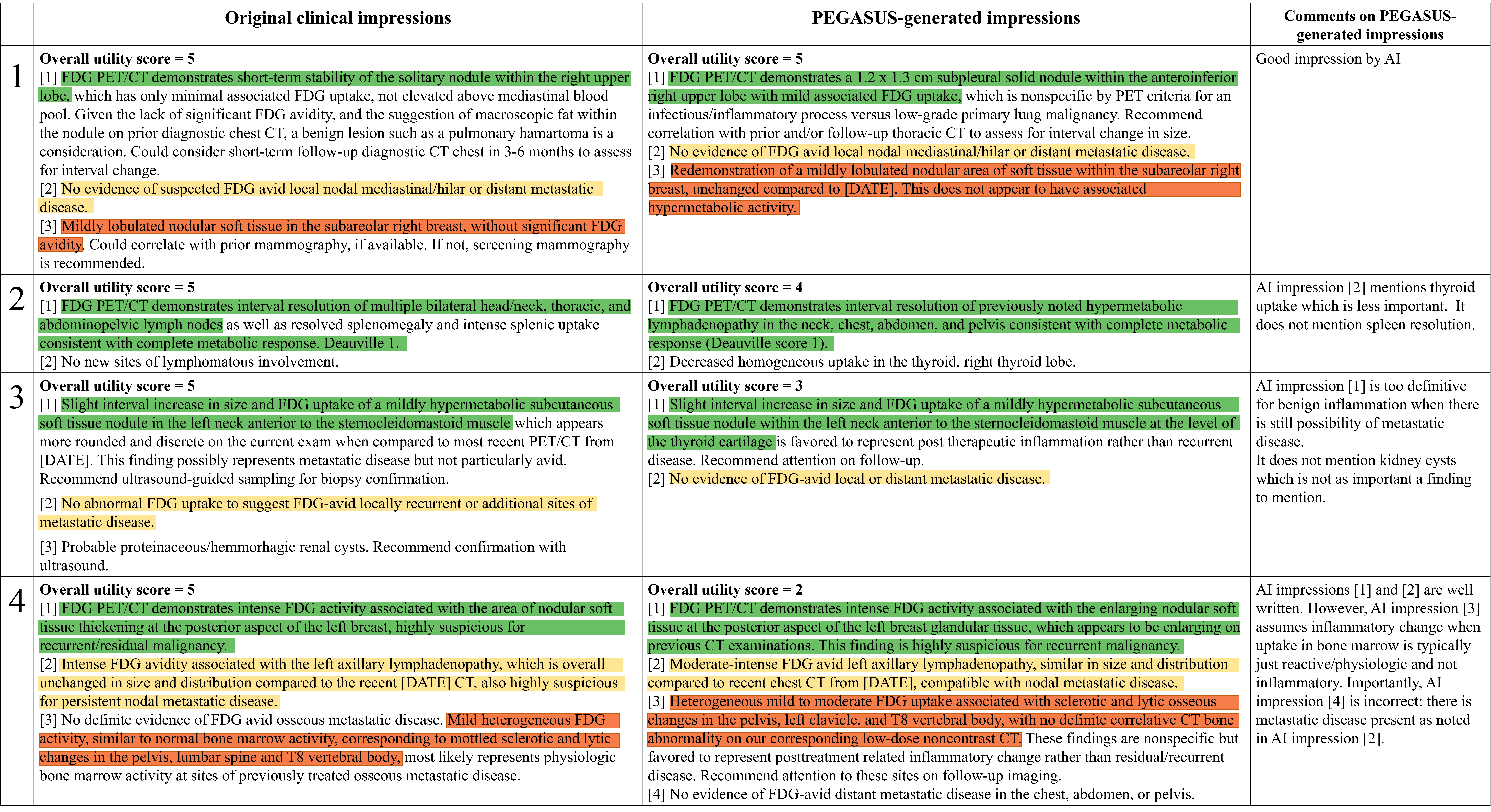}
  \caption{\small{A side-by-side comparison of clinical impressions and PEGASUS-generated impressions (overall utility scores range from 2 to 5). The last column presents comments from the physicians in our expert reader study. Sentences with similar semantic meanings in the original clinical impressions and the PEGASUS-generated impressions are highlighted using identical colors. Protected health information (PHI) has been anonymized and denoted with [X], where X may represent age or examination date.}} 
    \vspace{-2pt}
  \label{fig:fig5}
\end{figure}

\begin{figure}[h!]
\centering
\includegraphics[width=0.98\textwidth]{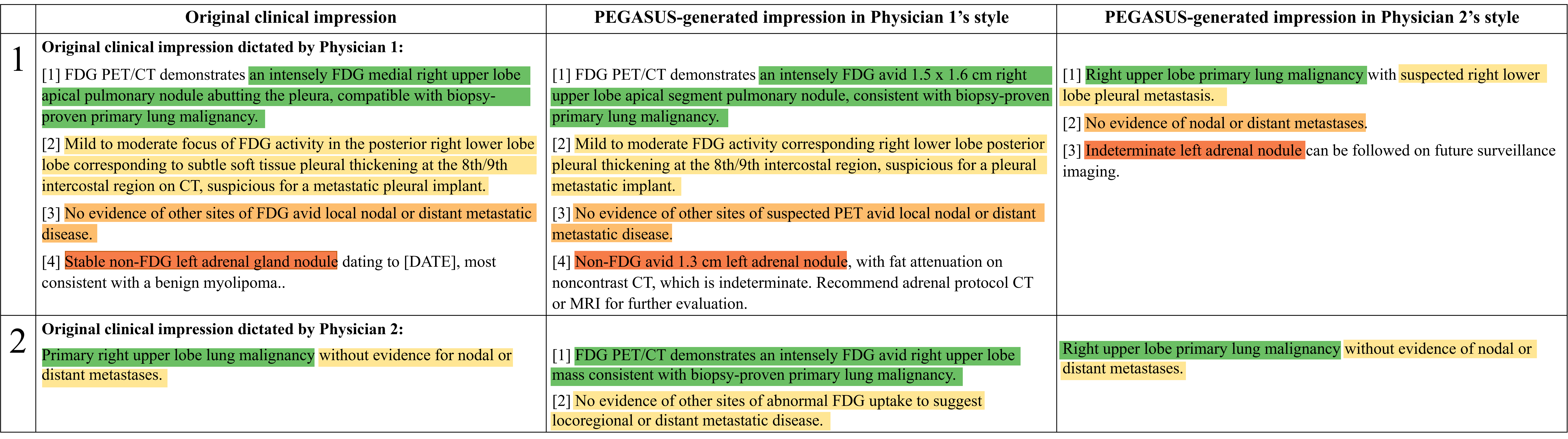}
  \caption{\small{Examples of PEGASUS-generated impressions customized for the physician’s reporting style. The first column shows the original clinical impressions: the first example from Physician 1 and the second from Physician 2. Subsequent columns present impressions generated in the style of Physician 1 and Physician 2, respectively. For each case, both impressions capture the critical findings and deliver similar diagnoses. However, their length, level of detail and phrasing generally reflect each physician’s unique style. Sentences with similar semantic meanings in the original clinical impressions and the PEGASUS-generated impressions are highlighted using identical colors.}} 
  \label{fig:fig6}
  \vspace{-10pt}
\end{figure}

\subsection{External Testing}
When PEGASUS was applied to the external test set, a significant drop ($P<$0.001) was observed in the evaluation metrics. Averaged across the reporting styles of Physicians 1, 2, and 3, BARTScore+PET in the external set was 15$\%$ worse than in the internal test set. Similarly, ROUGE-L decreased by 29$\%$ in the external set. Quantitative results are detailed in Appendix S10, along with four sample cases.

\section{Discussion}
In this work, we trained 12 language models on the task of PET impression generation. To identify the best metrics to evaluate model-generated impressions, we benchmarked 30 evaluation metrics against quality scores assigned by physicians and found that domain-adapted text-generation-based metrics, namely BARTScore+PET and PEGASUSScore+PET, exhibited the strongest correlation with physician preferences. These metrics selected PEGASUS as the top-performing LLM for our expert evaluation. A total of 72 cases were reviewed by three NM physicians, and the large majority of PEGASUS-generated impressions were rated as clinically acceptable. Moreover, by leveraging a specific token in the input to encode the reading physician’s identity, we enabled LLMs to learn different reporting styles and generate personalized impressions. When physicians assessed impressions generated in their own style, they considered these impressions to be of comparable overall utility to the impressions dictated by other physicians.

Past research on text summarization has introduced numerous evaluation metrics for assessing the quality of AI-generated summaries. However, when these metrics were employed to evaluate PET impressions, the majority did not align closely with physician judgments. This observation is consistent with findings from other works that evaluated medical document (33) or clinical note summarization (12). In general, we found that model-based metrics slightly outperformed lexical-based metrics, although better evaluation metrics are needed.

Based on our comparison of 12 language models, we observed that the biomedical-domain pretrained LLMs did not outperform their base models. This could be attributed to two reasons. First, our large training set diminished the benefits of medical-domain adaptation. Second, the corpora, such as MIMIC-III and PubMed, likely had limited PET related content, making pretraining less effective for our task. Additionally, we found that the large decoder-only models showed inferior performance in summarizing PET findings compared to the SOTA encoder-decoder models. It stems from their lack of an encoder mechanism that can efficiently distill the essence of input sequences. In this study, we did not test large proprietary models like GPT4 due to data ownership concerns and the inability to fine-tune the models for personalized impressions. Recent works (7, 8) explored their capability in radiology report summarization using the in-context learning technique. The question of whether this approach could surpass the full fine-tuning method for public LLMs and its suitability for clinical use remains to be answered. 
While most PEGASUS-generated impressions were deemed clinically acceptable in expert evaluation, it is crucial to understand what mistakes are commonly committed by the LLM. First, the main problem in model-generated impressions is factual inaccuracies, which manifest as misinterpretation of findings or contradictory statements. Second, the diagnoses given by the LLM could sometimes be overly definite without adequate supporting evidence. Third, some recommendations for clinical follow-up were non-specific, offering limited guidance for patient management. It is worth mentioning that final diagnoses and recommendations are usually not included in the report findings and must be inferred by the model. These observations underscore the need for review and appropriate editing by physicians before report finalization. Of note, LLM-based impression generation can be akin to preliminary impression drafts by radiology resident trainees provided for review by the radiology faculty in an academic training setting.

This study had several limitations. First, when fine-tuning LLaMA and Alpaca, we only investigated a lightweight domain adaptation method, LoRA, constrained by computational resources. Second, we controlled the style of generated impressions by altering a specific token in the input, leaving other potential techniques unexplored. Third, during external testing, we observed a moderate decrease in the evaluation metrics. This is expected given the differences in reporting styles between our internal and external physicians. However, whether this result aligns with physician judgments remains uncertain and warrants further investigation. Lastly, our training dataset was restricted to a single institution. Future work should be expanding our research to a multi-center study.

To conclude, we systematically investigated the potential of LLMs to automate impression generation for whole-body PET reports. Our reader study showed that the top-performing LLM, PEGASUS, produced clinically useful and personalized impressions for the majority of cases. Given its performance, we believe our model could be integrated into clinical workflows and expedite PET reporting by automatically drafting initial impressions based on the findings. 

\section*{Acknowledgments}

We acknowledge funding support from Imaging and Radiology Oncology Core Rhode Island (U24CA180803), Biomarker, Imaging and Quality of Life Studies Funding Program (BIQSFP), NCTN Operations Center Grant U10CA180886, NCTN Statistics $\&$ Data Center Grant U10CA180899 and St. Baldrick's Foundation.

Disclaimer: The content is solely the responsibility of the authors and does not necessarily represent the official views of the National Institutes of Health.

%Bibliography
\bibliographystyle{unsrt}  
\bibliography{references}  
\begin{itemize}[leftmargin=*]
\item[1.]  Niederkohr RD, Greenspan BS, Prior JO, et al. Reporting Guidance for Oncologic 18 F-FDG PET/CT Imaging. J Nucl Med. 2013;54(5):756–761. doi: \url{http://doi.org/10.2967/jnumed.112.112177}.
\item[2.]  Hartung MP, Bickle IC, Gaillard F, Kanne JP. How to Create a Great Radiology Report. RadioGraphics. 2020;40(6):1658–1670. doi: \url{http://doi.org/10.1148/rg.2020200020}.
\item[3.]  Zhang Y, Ding DY, Qian T, Manning CD, Langlotz CP. Learning to Summarize Radiology Findings. arXiv; 2018. \url{http://arxiv.org/abs/1809.04698}. Accessed March 1, 2023.
\item[4.]  Hu J, Li Z, Chen Z, Li Z, Wan X, Chang T-H. Graph Enhanced Contrastive Learning for Radiology Findings Summarization. arXiv; 2022. \url{http://arxiv.org/abs/2204.00203}. Accessed March 2, 2023.
\item[5.]  Delbrouck J-B, Varma M, Langlotz CP. Toward expanding the scope of radiology report summarization to multiple anatomies and modalities. arXiv; 2022. \url{http://arxiv.org/abs/2211.08584}. Accessed March 2, 2023.
\item[6.]  Liu Z, Zhong A, Li Y, et al. Radiology-GPT: A Large Language Model for Radiology. arXiv; 2023. \url{http://arxiv.org/abs/2306.08666}. Accessed July 20, 2023.
\item[7.]  Sun Z, Ong H, Kennedy P, et al. Evaluating GPT4 on Impressions Generation in Radiology Reports. Radiology. 2023;307(5):e231259. doi: \url{http://doi.org/10.1148/radiol.231259}.
\item[8.]  Ma C, Wu Z, Wang J, et al. ImpressionGPT: An Iterative Optimizing Framework for Radiology Report Summarization with ChatGPT. arXiv; 2023. \url{http://arxiv.org/abs/2304.08448}. Accessed August 14, 2023.
\item[9.]  Johnson AEW, Pollard TJ, Berkowitz SJ, et al. MIMIC-CXR, a de-identified publicly available database of chest radiographs with free-text reports. Sci Data. 2019;6(1):317. doi: \url{http://doi.org/10.1038/s41597-019-0322-0}.
\item[10.]  Hu J, Li J, Chen Z, et al. Word Graph Guided Summarization for Radiology Findings. arXiv; 2021. \url{http://arxiv.org/abs/2112.09925}. Accessed August 22, 2023.
\item[11.]  Smit A, Jain S, Rajpurkar P, Pareek A, Ng AY, Lungren MP. CheXbert: Combining Automatic Labelers and Expert Annotations for Accurate Radiology Report Labeling Using BERT. arXiv; 2020. \url{http://arxiv.org/abs/2004.09167}. Accessed August 27, 2023.
\item[12.]  Abacha AB, Yim W, Michalopoulos G, Lin T. An Investigation of Evaluation Metrics for Automated Medical Note Generation. arXiv; 2023. \url{http://arxiv.org/abs/2305.17364}. Accessed August 27, 2023.
\item[13.]  Kayaalp M, Browne AC, Dodd ZA, Sagan P, McDonald CJ. De-identification of Address, Date, and Alphanumeric Identifiers in Narrative Clinical Reports. AMIA Annu Symp Proc; 2014; 2014: 767–776. PMID: 25954383; PMCID: PMC4419982. 
\item[14.]  Castellino SM, Pei Q, Parsons SK, et al. Brentuximab Vedotin with Chemotherapy in Pediatric High-Risk Hodgkin’s Lymphoma. N Engl J Med. 2022;387(18):1649–1660. doi: \url{http://doi.org/10.1056/NEJMoa2206660}.
\item[15.]  Wang Y, Kordi Y, Mishra S, et al. Self-Instruct: Aligning Language Models with Self-Generated Instructions. arXiv; 2023. \url{http://arxiv.org/abs/2212.10560}. Accessed June 20, 2023.
\item[16.]  Rohan T, Ishaan G, Tianyi Z, et al. Stanford Alpaca: An Instruction-following LLaMA model. GitHub; 2023. \url{https://github.com/tatsu-lab/stanford_alpaca}. Accessed June 20, 2023.
\item[17.] Lewis M, Liu Y, Goyal N, et al. BART: Denoising Sequence-to-Sequence Pre-training for Natural Language Generation, Translation, and Comprehension. arXiv; 2019. \url{http://arxiv.org/abs/1910.13461}. Accessed March 7, 2023.
\item[18.]  Zhang J, Zhao Y, Saleh M, Liu PJ. PEGASUS: Pre-training with Extracted Gap-sentences for Abstractive Summarization. arXiv; 2020. \url{http://arxiv.org/abs/1912.08777}. Accessed March 7, 2023.
\item[19.]  Raffel C, Shazeer N, Roberts A, et al. Exploring the Limits of Transfer Learning with a Unified Text-to-Text Transformer. arXiv; 2020. \url{http://arxiv.org/abs/1910.10683}. Accessed June 20, 2023.
\item[20.]  Wei J, Bosma M, Zhao VY, et al. Finetuned Language Models Are Zero-Shot Learners. arXiv; 2022. \url{http://arxiv.org/abs/2109.01652}. Accessed August 15, 2023.
\item[21.]  Yuan H, Yuan Z, Gan R, Zhang J, Xie Y, Yu S. BioBART: Pretraining and Evaluation of A Biomedical Generative Language Model. arXiv; 2022. \url{http://arxiv.org/abs/2204.03905}. Accessed August 15, 2023.
\item[22.]  Lu Q, Dou D, Nguyen TH. ClinicalT5: A Generative Language Model for Clinical Text. Findings of the Association for Computational Linguistics: EMNLP 2022, pages 5436–5443, Abu Dhabi, United Arab Emirates. Association for Computational Linguistics. doi: \url{http://doi.org/10.18653/v1/2022.findings-emnlp.398}.
\item[23.]  Chen C, Yin Y, Shang L, et al. bert2BERT: Towards Reusable Pretrained Language Models. Proceedings of the 60th Annual Meeting of the Association for Computational Linguistics (Volume 1: Long Papers). Dublin, Ireland: Association for Computational Linguistics; 2022. p. 2134–2148. doi: \url{http://doi.org/10.18653/v1/2022.acl-long.151}.
\item[24.]  Ziegler DM, Stiennon N, Wu J, et al. Fine-Tuning Language Models from Human Preferences. arXiv; 2020. \url{http://arxiv.org/abs/1909.08593}. Accessed June 20, 2023.
\item[25.]  Zhang S, Roller S, Goyal N, et al. OPT: Open Pre-trained Transformer Language Models. arXiv; 2022. \url{http://arxiv.org/abs/2205.01068}. Accessed August 15, 2023.
\item[26.]  Touvron H, Lavril T, Izacard G, et al. LLaMA: Open and Efficient Foundation Language Models. arXiv; 2023. \url{http://arxiv.org/abs/2302.13971}. Accessed August 14, 2023.
\item[27.]  Hu EJ, Shen Y, Wallis P, et al. LoRA: Low-Rank Adaptation of Large Language Models. arXiv; 2021. \url{http://arxiv.org/abs/2106.09685}. Accessed August 15, 2023.
\item[28.]  Yuan W, Neubig G, Liu P. BARTScore: Evaluating Generated Text as Text Generation. arXiv; 2021. \url{http://arxiv.org/abs/2106.11520}. Accessed August 15, 2023.
\item[29.]  Huemann Z, Lee C, Hu J, Cho SY, Bradshaw T. Domain-adapted large language models for classifying nuclear medicine reports. arXiv; 2023. \url{http://arxiv.org/abs/2303.01258}. Accessed March 17, 2023.
\item[30.]  Smith L, Tanabe LK, Ando RJ nee, et al. Overview of BioCreative II gene mention recognition. Genome Biol. 2008;9(S2):S2. doi: \url{http://doi.org/10.1186/gb-2008-9-s2-s2}.
\item[31.]  Lin CY. ROUGE: A package for automatic evaluation of summaries. In Text Summarization Branches Out, Barcelona, Spain, July 2004. Association for Computational Linguistics, 2004; 74–81. \url{https://aclanthology.org/W04-1013/}. 
\item[32.]  Zhang T, Kishore V, Wu F, Weinberger KQ, Artzi Y. BERTScore: Evaluating Text Generation with BERT. arXiv; 2020. \url{http://arxiv.org/abs/1904.09675}. Accessed August 22, 2023.
\item[33.]  Wang LL, Otmakhova Y, DeYoung J, et al. Automated Metrics for Medical Multi-Document Summarization Disagree with Human Evaluations. arXiv; 2023. \url{http://arxiv.org/abs/2305.13693}.  Accessed August 22, 2023. 
\end{itemize}

\begin{center}
\begin{figure}[h!]
\vspace{-90pt}
\hspace{-60pt}
\includegraphics[width=1.25\textwidth]{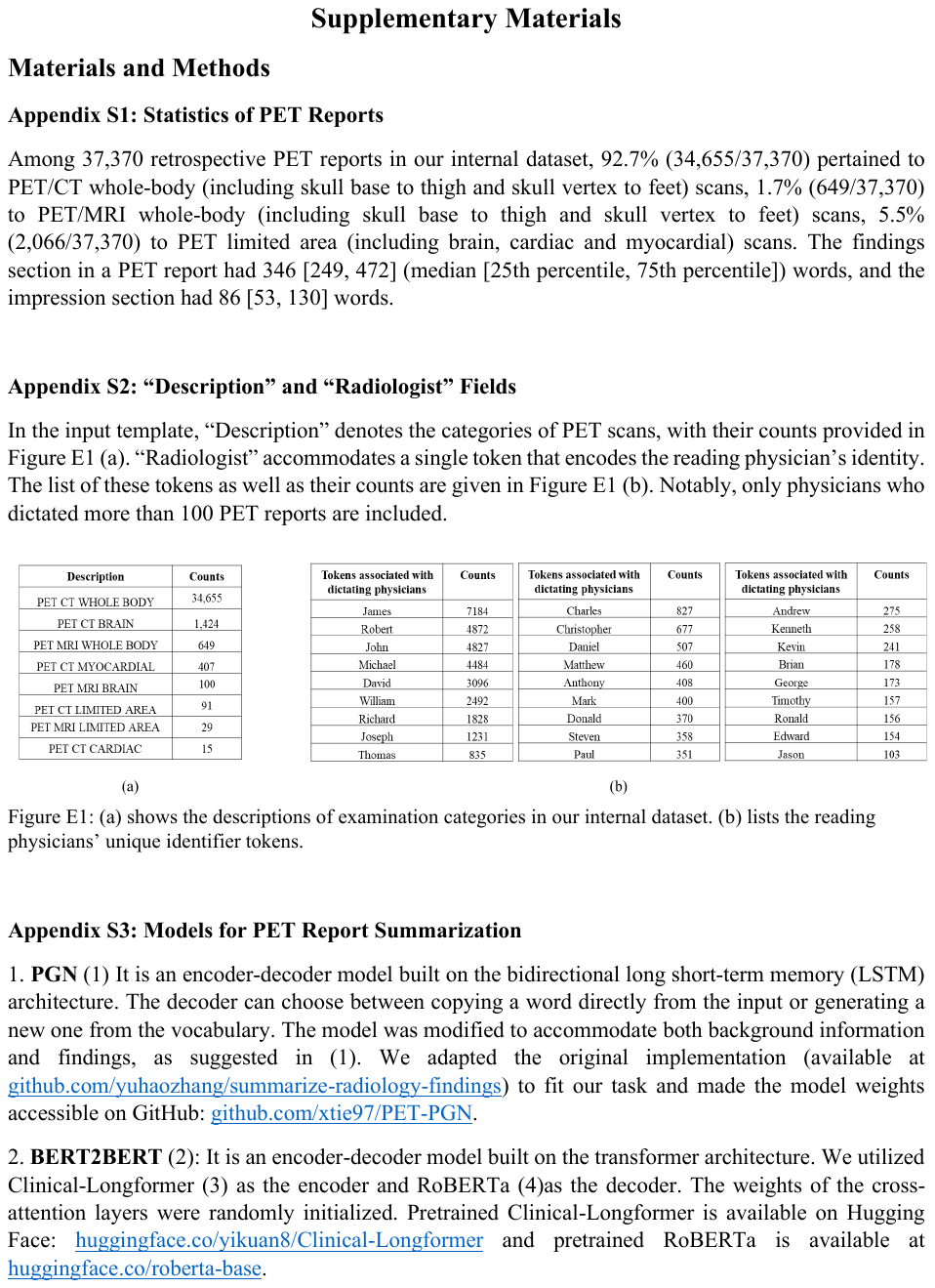}
\end{figure}
\end{center}

\begin{center}
\begin{figure}[h!]
\vspace{-90pt}
\hspace{-60pt}
\includegraphics[width=1.25\textwidth]{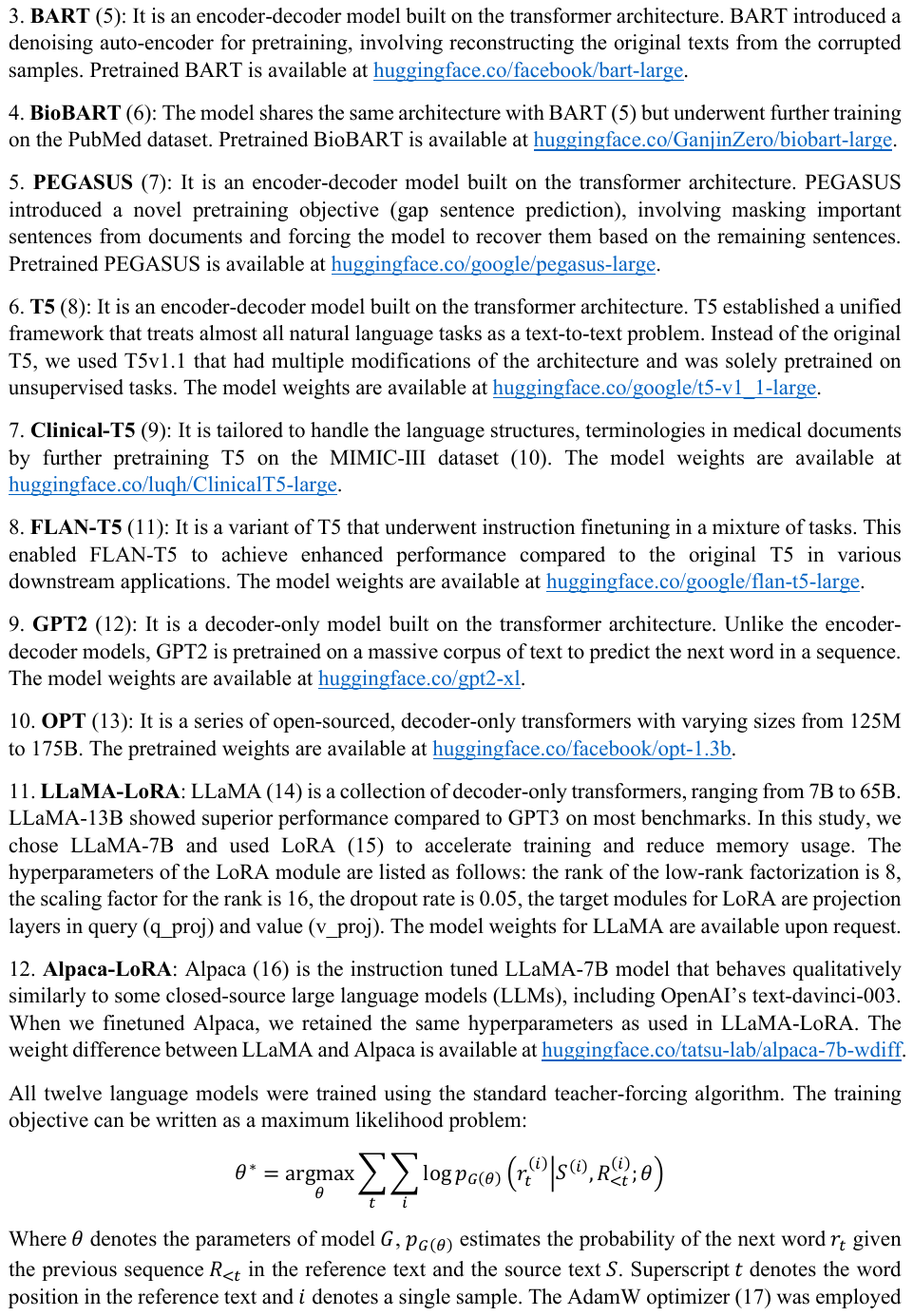}
\end{figure}
\end{center}

\begin{center}
\begin{figure}[h!]
\vspace{-90pt}
\hspace{-60pt}
\includegraphics[width=1.25\textwidth]{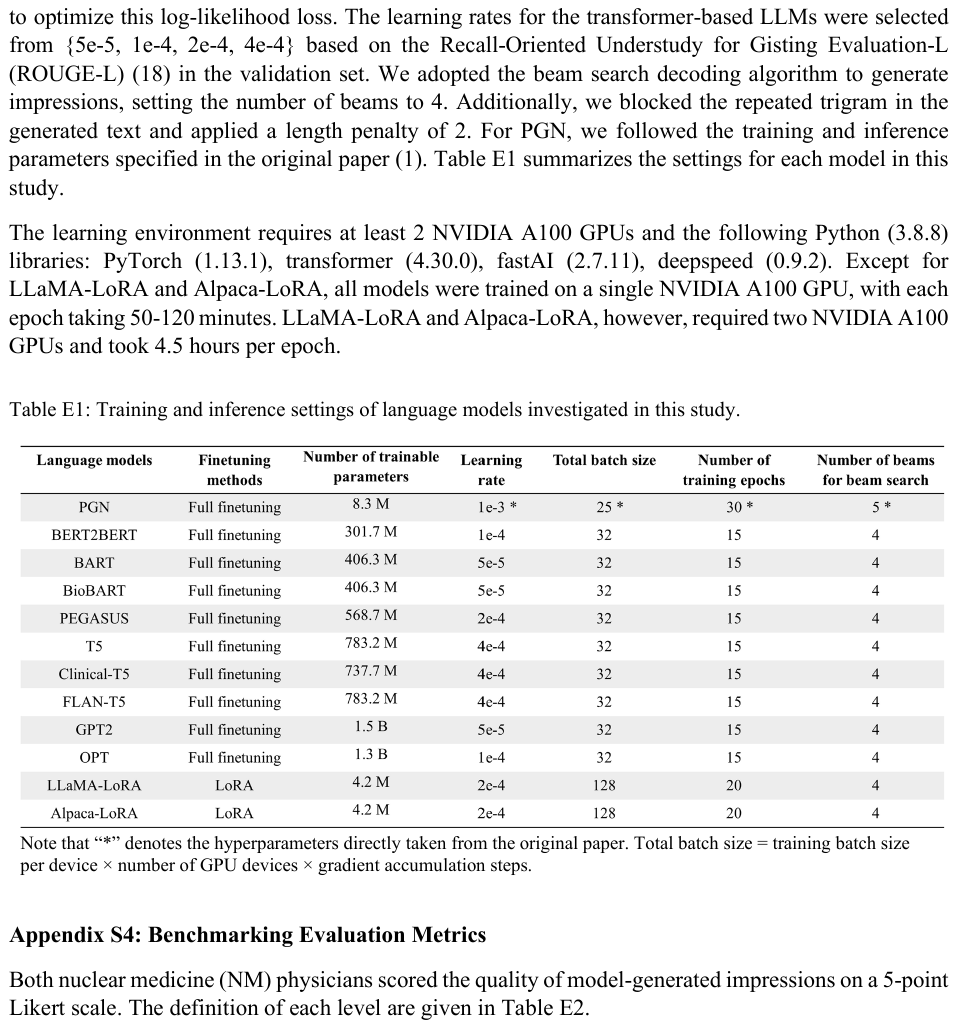}
\end{figure}
\end{center}

\begin{center}
\begin{figure}[h!]
\vspace{-90pt}
\hspace{-60pt}
\includegraphics[width=1.25\textwidth]{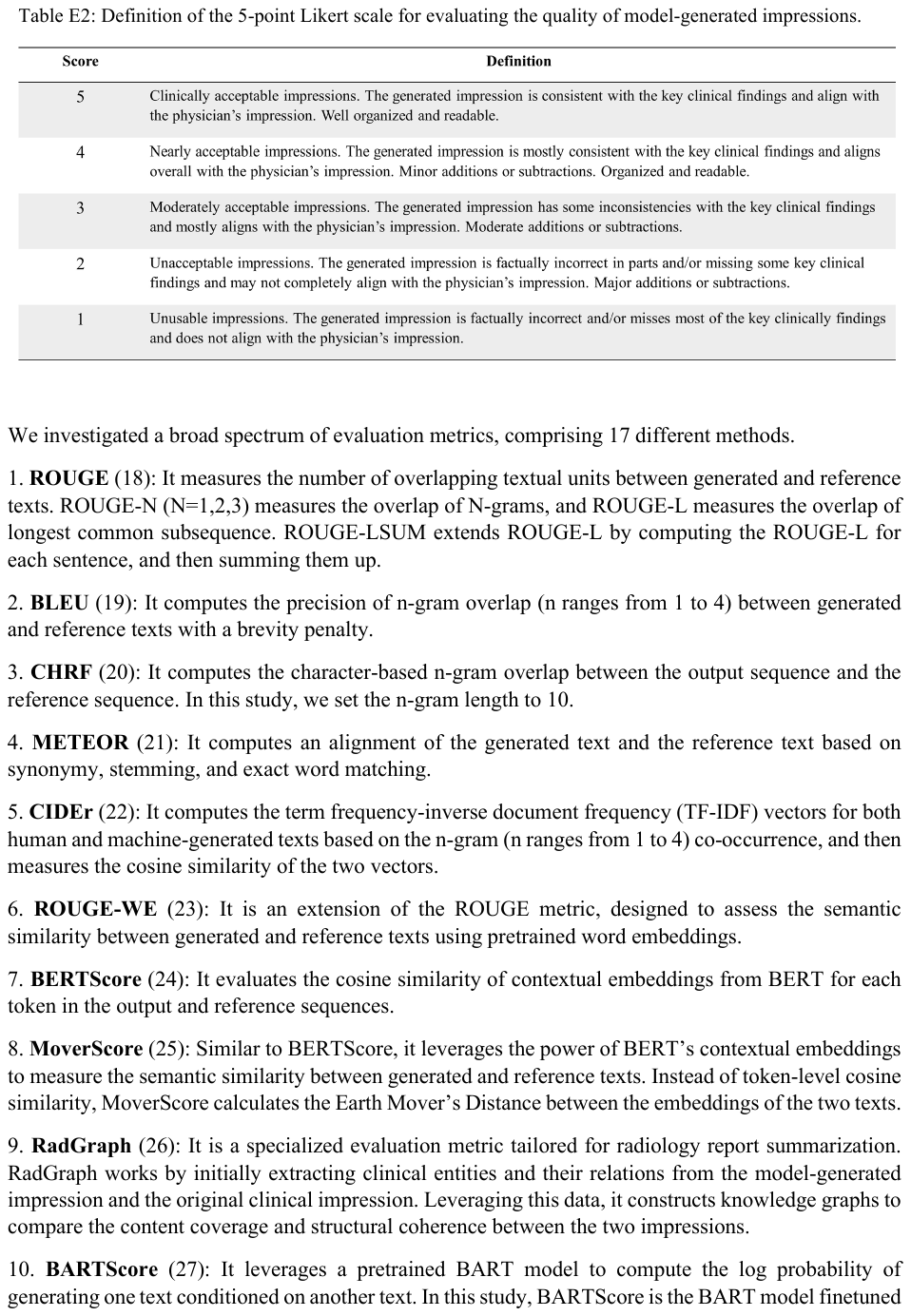}
\end{figure}
\end{center}

\begin{center}
\begin{figure}[h!]
\vspace{-90pt}
\hspace{-60pt}
\includegraphics[width=1.25\textwidth]{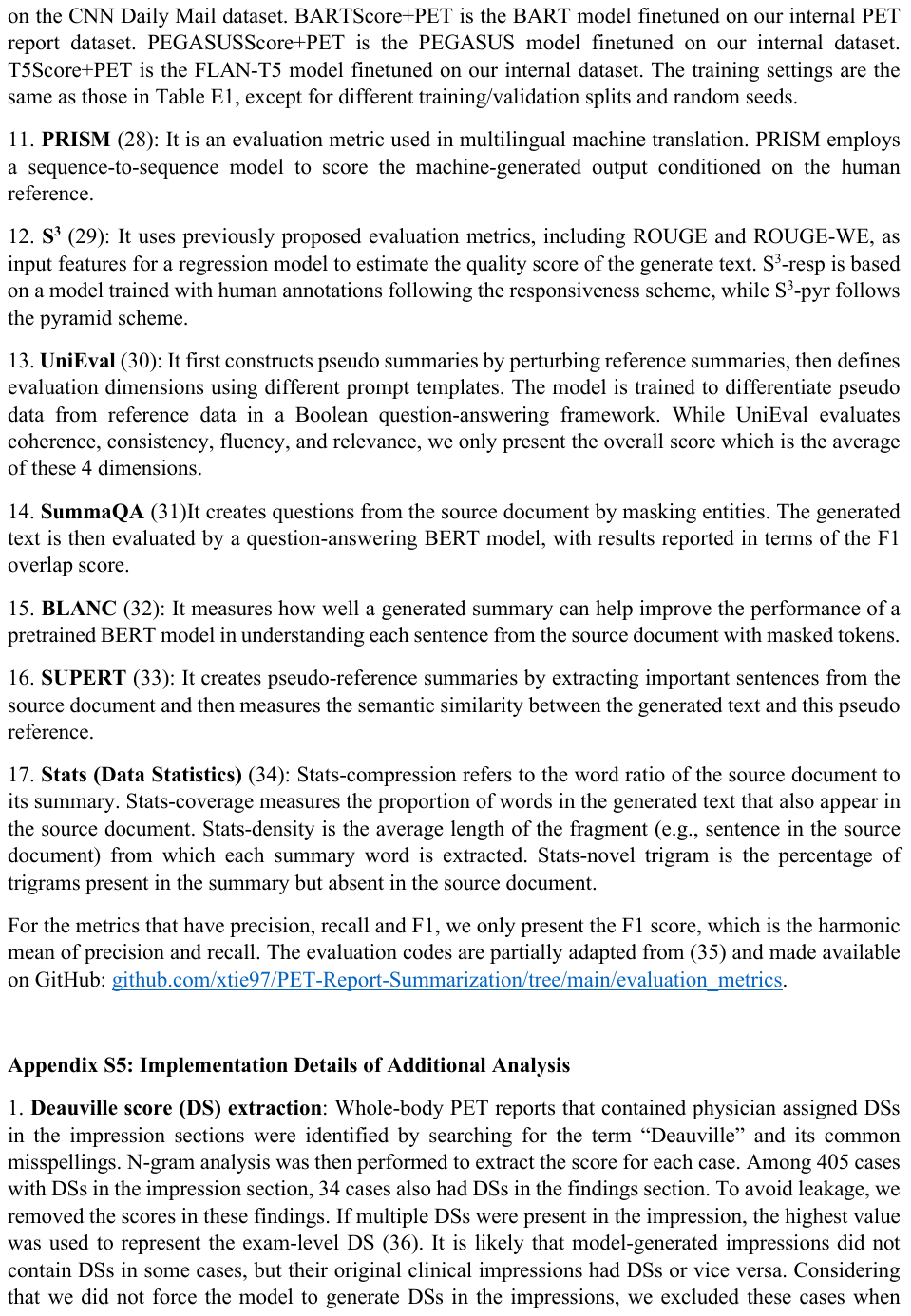}
\end{figure}
\end{center}

\begin{center}
\begin{figure}[h!]
\vspace{-90pt}
\hspace{-60pt}
\includegraphics[width=1.25\textwidth]{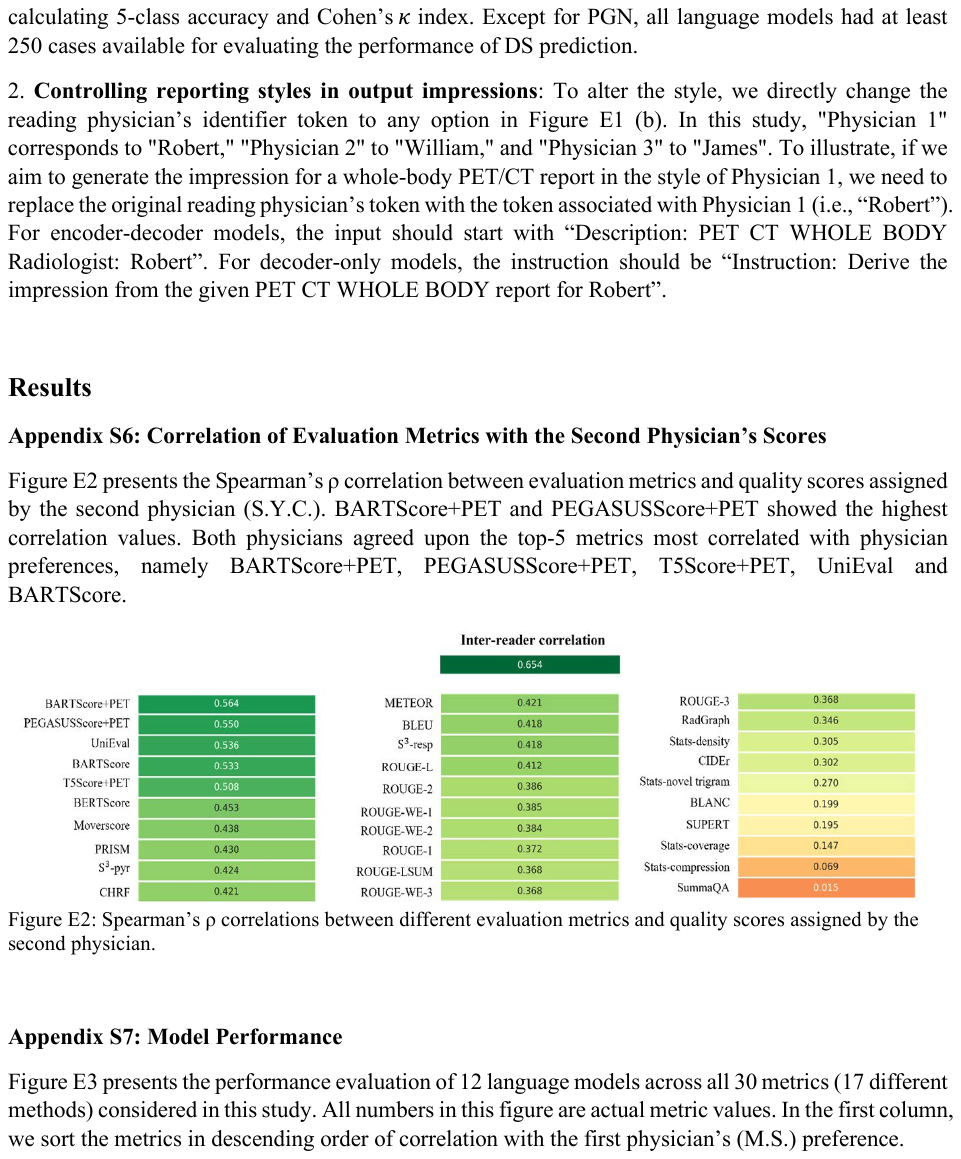}
\end{figure}
\end{center}

\begin{center}
\begin{figure}[h!]
\vspace{-90pt}
\hspace{-60pt}
\includegraphics[width=1.25\textwidth]{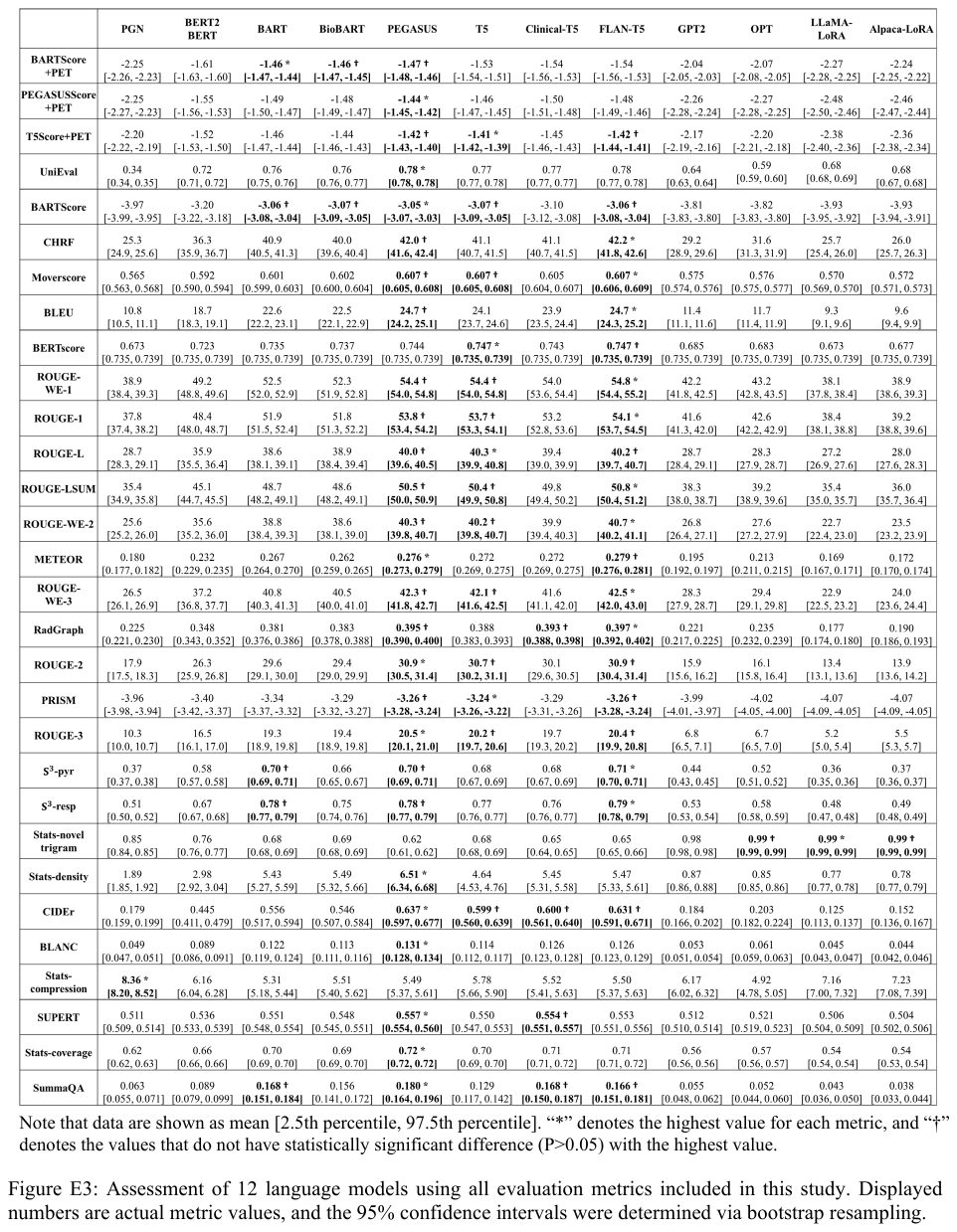}
\end{figure}
\end{center}

\begin{center}
\begin{figure}[h!]
\vspace{-90pt}
\hspace{-60pt}
\includegraphics[width=1.25\textwidth]{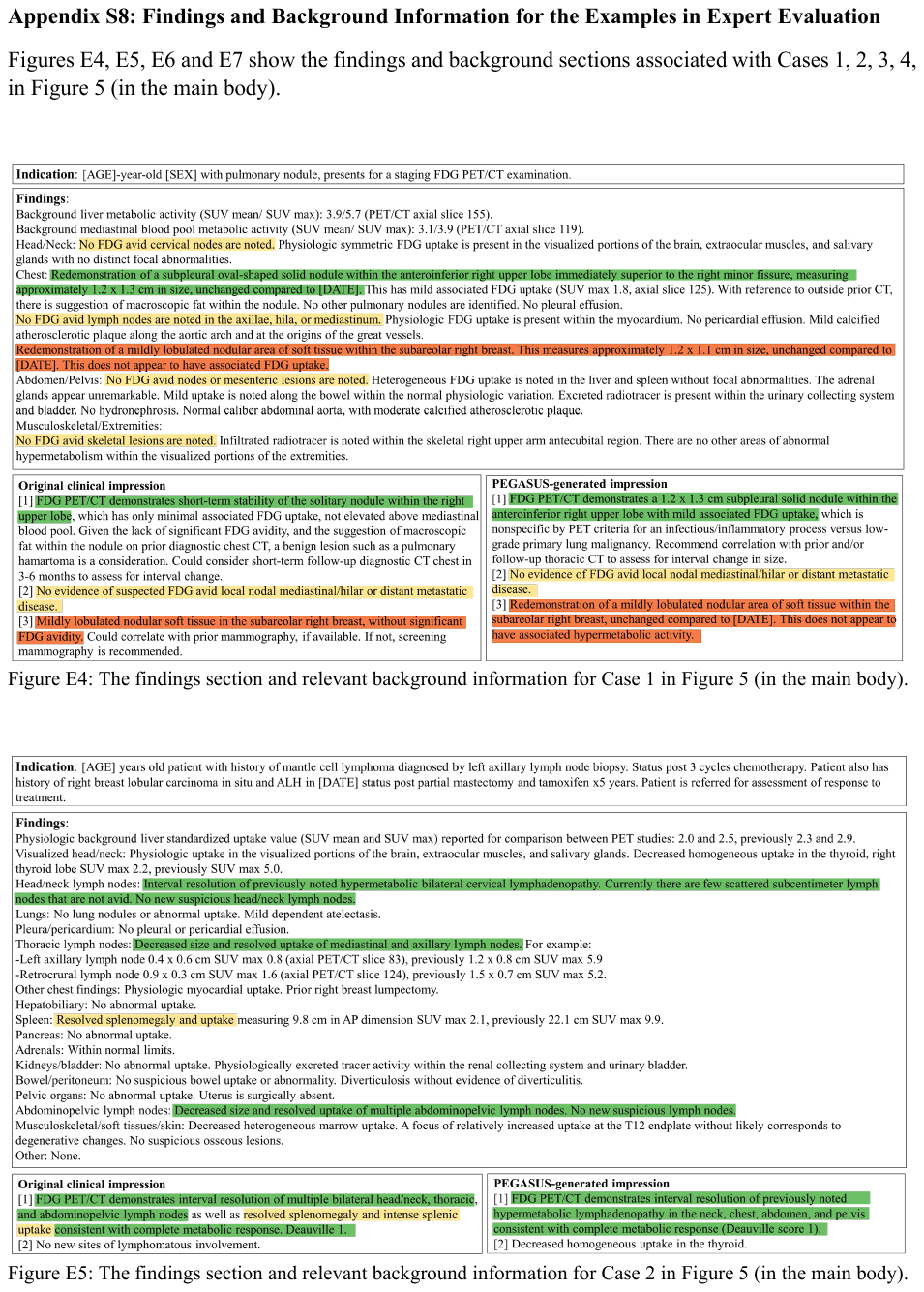}
\end{figure}
\end{center}

\begin{center}
\begin{figure}[h!]
\vspace{-90pt}
\hspace{-60pt}
\includegraphics[width=1.25\textwidth]{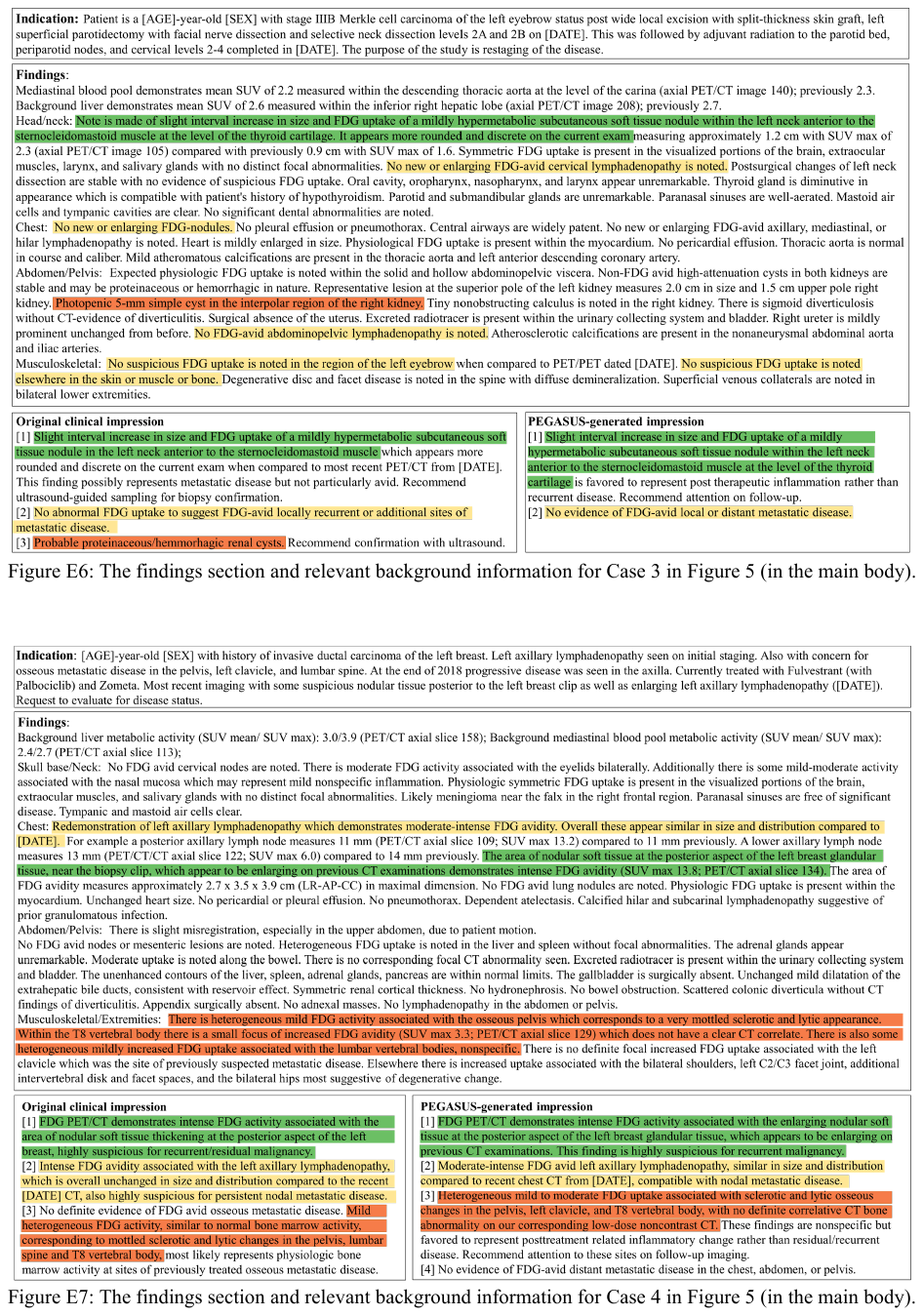}
\end{figure}
\end{center}

\begin{center}
\begin{figure}[h!]
\vspace{-90pt}
\hspace{-60pt}
\includegraphics[width=1.25\textwidth]{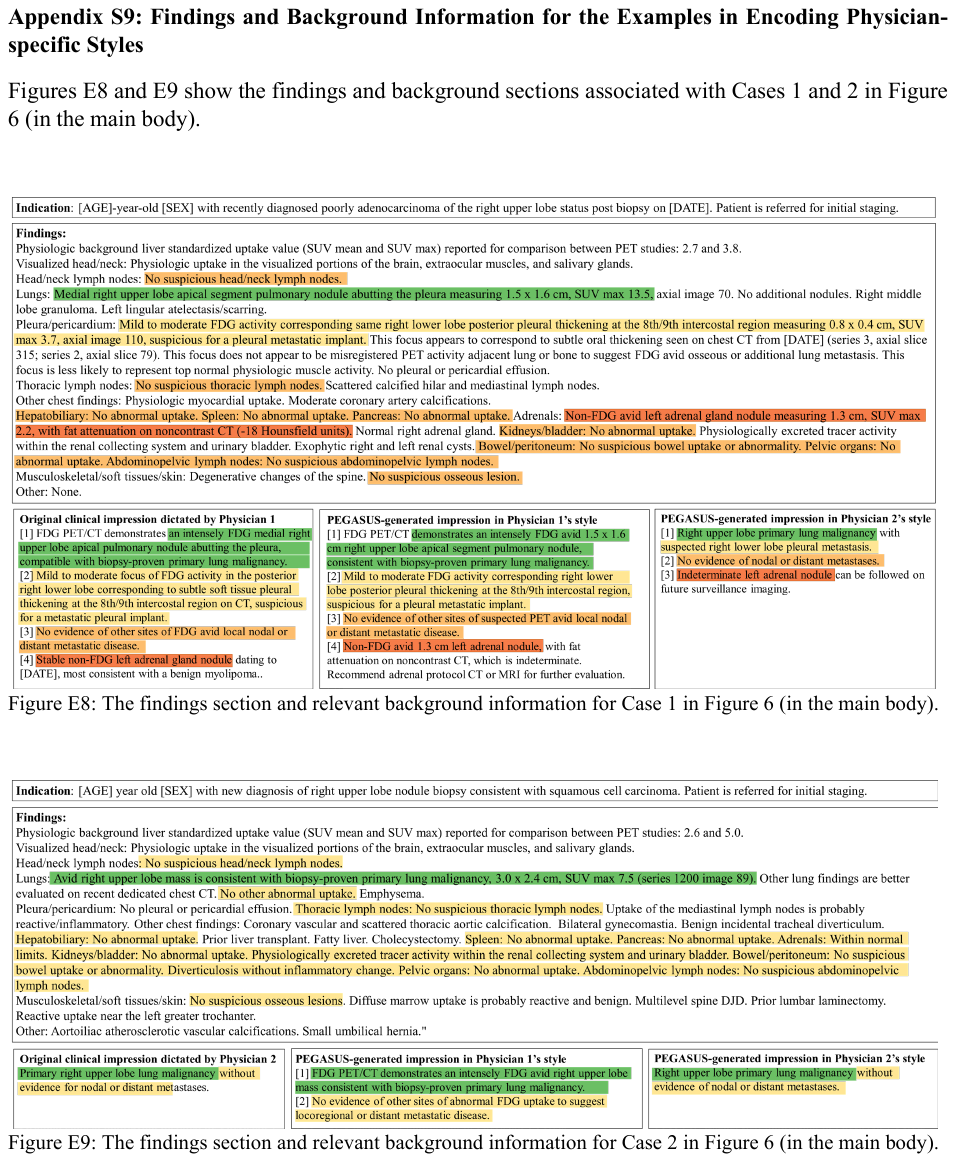}
\end{figure}
\end{center}

\begin{center}
\begin{figure}[h!]
\vspace{-90pt}
\hspace{-60pt}
\includegraphics[width=1.25\textwidth]{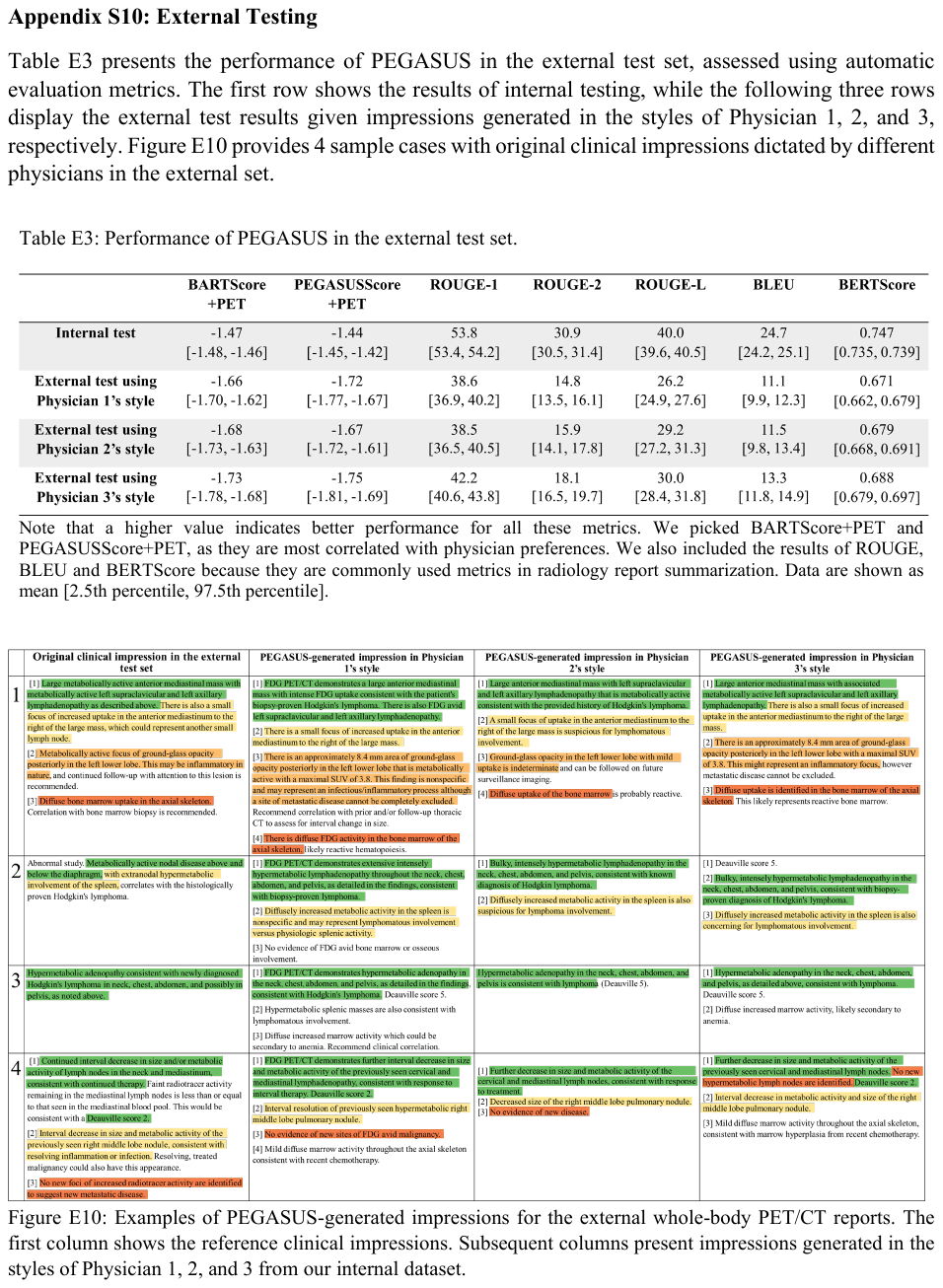}
\end{figure}
\end{center}

\clearpage
\section*{References}
\begin{itemize}[leftmargin=*]
\item [1.] Zhang Y, Ding DY, Qian T, Manning CD, Langlotz CP. Learning to Summarize Radiology Findings. Proc Ninth Int Workshop Health Text Min Inf Anal. Brussels, Belgium: Association for Computational Linguistics; 2018. p. 204–213. doi: \url{http://doi.org/10.18653/v1/W18-5623}.

\item [2.] 	Chen C, Yin Y, Shang L, et al. bert2BERT: Towards Reusable Pretrained Language Models. Proc 60th Annu Meet Assoc Comput Linguist Vol 1 Long Pap. Dublin, Ireland: Association for Computational Linguistics; 2022. p. 2134–2148. doi: \url{http://doi.org/10.18653/v1/2022.acl-long.151}.

\item [3.] 	Li Y, Wehbe RM, Ahmad FS, Wang H, Luo Y. Clinical-Longformer and Clinical-BigBird: Transformers for long clinical sequences. arXiv; 2022. \url{http://arxiv.org/abs/2201.11838}. Accessed August 16, 2023.

\item [4.] 	Liu Y, Ott M, Goyal N, et al. RoBERTa: A Robustly Optimized BERT Pretraining Approach. arXiv; 2019.  \url{http://arxiv.org/abs/1907.11692}. Accessed August 16, 2023.

\item [5.] 	Lewis M, Liu Y, Goyal N, et al. BART: Denoising Sequence-to-Sequence Pre-training for Natural Language Generation, Translation, and Comprehension. arXiv; 2019.  \url{http://arxiv.org/abs/1910.13461}. Accessed March 7, 2023.

\item [6.] 	Yuan H, Yuan Z, Gan R, Zhang J, Xie Y, Yu S. BioBART: Pretraining and Evaluation of A Biomedical Generative Language Model. arXiv; 2022.  \url{http://arxiv.org/abs/2204.03905}. Accessed August 15, 2023.

\item [7.] 	Zhang J, Zhao Y, Saleh M, Liu PJ. PEGASUS: Pre-training with Extracted Gap-sentences for Abstractive Summarization. arXiv; 2020.  \url{http://arxiv.org/abs/1912.08777}. Accessed March 7, 2023.

\item [8.] 	Raffel C, Shazeer N, Roberts A, et al. Exploring the Limits of Transfer Learning with a Unified Text-to-Text Transformer. arXiv; 2020.  \url{http://arxiv.org/abs/1910.10683}. Accessed August 14, 2023.

\item [9.] 	Lu Q, Dou D, Nguyen TH. ClinicalT5: A Generative Language Model for Clinical Text. Findings of the Association for Computational Linguistics: EMNLP 2022, pages 5436–5443, Abu Dhabi, United Arab Emirates. Association for Computational Linguistics. doi: \url{http://doi.org/10.18653/v1/2022.findings-emnlp.398}.

\item [10.] 	Johnson AEW, Pollard TJ, Berkowitz SJ, et al. MIMIC-CXR, a de-identified publicly available database of chest radiographs with free-text reports. Sci Data. 2019;6(1):317. doi: \url{http://doi.org/10.1038/s41597-019-0322-0}.

\item [11.] 	Wei J, Bosma M, Zhao VY, et al. Finetuned Language Models Are Zero-Shot Learners. arXiv; 2022.  \url{http://arxiv.org/abs/2109.01652}. Accessed August 15, 2023.

\item [12.] 	Ziegler DM, Stiennon N, Wu J, et al. Fine-Tuning Language Models from Human Preferences. arXiv; 2020.  \url{http://arxiv.org/abs/1909.08593}. Accessed August 14, 2023.

\item [13.] 	Zhang S, Roller S, Goyal N, et al. OPT: Open Pre-trained Transformer Language Models. arXiv; 2022.  \url{http://arxiv.org/abs/2205.01068}. Accessed February 22, 2023.

\item [14.] 	Touvron H, Lavril T, Izacard G, et al. LLaMA: Open and Efficient Foundation Language Models. arXiv; 2023.  \url{http://arxiv.org/abs/2302.13971}. Accessed August 14, 2023.

\item [15.] 	Hu EJ, Shen Y, Wallis P, et al. LoRA: Low-Rank Adaptation of Large Language Models. arXiv; 2021.  \url{http://arxiv.org/abs/2106.09685}. Accessed August 15, 2023.

\item [16.]  Taori R, Gulrajani I, Zhang T, et al. Stanford Alpaca: An Instruction-following LLaMA model. GitHub; 2023.  \url{https://github.com/tatsu-lab/stanford_alpaca}. Accessed June 20, 2023.

\item [17.] 	Loshchilov I, Hutter F. Decoupled Weight Decay Regularization. arXiv; 2019.  \url{http://arxiv.org/abs/1711.05101}. Accessed August 31, 2023.

\item [18.] 	Lin CY. ROUGE: A package for automatic evaluation of summaries. In Text Summarization Branches Out, Barcelona, Spain, July 2004. Association for Computational Linguistics, 2004; 74–81.  \url{https://aclanthology.org/W04-1013/}.

\item [19.] 	Papineni K, Roukos S, Ward T, Zhu W-J. BLEU: a method for automatic evaluation of machine translation. Proc 40th Annu Meet Assoc Comput Linguist - ACL 02. Philadelphia, Pennsylvania: Association for Computational Linguistics; 2001. p. 311. doi: \url{http://doi.org/10.3115/1073083.1073135}.

\item [20.] 	Popović M. chrF: character n-gram F-score for automatic MT evaluation. Proc Tenth Workshop Stat Mach Transl. Lisbon, Portugal: Association for Computational Linguistics; 2015. p. 392–395. doi: \url{http://doi.org/10.18653/v1/W15-3049}.

\item [21.] 	Banerjee, S. and Lavie, A. METEOR: An Automatic Metric for MT Evaluation with Improved Correlation with Human Judgments. Proceedings of Workshop on Intrinsic and Extrinsic Evaluation Measures for MT and/or Summarization. Ann Arbor, Michigan: Association of Computational Linguistics, 2005. p. 65–72.

\item [22.] 	Vedantam R, Zitnick CL, Parikh D. CIDEr: Consensus-based Image Description Evaluation. arXiv; 2015.  \url{http://arxiv.org/abs/1411.5726}. Accessed August 31, 2023.

\item [23.] 	Ng J-P, Abrecht V. Better Summarization Evaluation with Word Embeddings for ROUGE. arXiv; 2015.  \url{http://arxiv.org/abs/1508.06034}. Accessed August 31, 2023.

\item [24.] 	Zhang T, Kishore V, Wu F, Weinberger KQ, Artzi Y. BERTScore: Evaluating Text Generation with BERT. arXiv; 2020.  \url{http://arxiv.org/abs/1904.09675}. Accessed August 22, 2023.

\item [25.] 	Zhao W, Peyrard M, Liu F, Gao Y, Meyer CM, Eger S. MoverScore: Text Generation Evaluating with Contextualized Embeddings and Earth Mover Distance. Proc 2019 Conf Empir Methods Nat Lang Process 9th Int Jt Conf Nat Lang Process EMNLP-IJCNLP. Hong Kong, China: Association for Computational Linguistics; 2019. p. 563–578. doi: \url{http://doi.org/10.18653/v1/D19-1053}.

\item [26.] 	Hu J, Li J, Chen Z, et al. Word Graph Guided Summarization for Radiology Findings. arXiv; 2021. \url{http://arxiv.org/abs/2112.09925}. Accessed March 2, 2023.

\item [27.] 	Yuan W, Neubig G, Liu P. BARTScore: Evaluating Generated Text as Text Generation. arXiv; 2021. \url{http://arxiv.org/abs/2106.11520}. Accessed August 15, 2023.

\item [28.] 	Thompson B, Post M. Automatic Machine Translation Evaluation in Many Languages via Zero-Shot Paraphrasing. Proc 2020 Conf Empir Methods Nat Lang Process EMNLP. Online: Association for Computational Linguistics; 2020. p. 90–121. doi: \url{http://doi.org/10.18653/v1/2020.emnlp-main.8}.

\item [29.] 	Peyrard M, Botschen T, Gurevych I. Learning to Score System Summaries for Better Content Selection Evaluation. Proc Workshop New Front Summ. Copenhagen, Denmark: Association for Computational Linguistics; 2017. p. 74–84. doi: \url{http://doi.org/10.18653/v1/W17-4510}.

\item [30.] 	Zhong M, Liu Y, Yin D, et al. Towards a Unified Multi-Dimensional Evaluator for Text Generation. Proc 2022 Conf Empir Methods Nat Lang Process. Abu Dhabi, United Arab Emirates: Association for Computational Linguistics; 2022. p. 2023–2038. doi: \url{http://doi.org/10.18653/v1/2022.emnlp-main.131}.

\item [31.] 	Scialom T, Lamprier S, Piwowarski B, Staiano J. Answers Unite! Unsupervised Metrics for Reinforced Summarization Models. arXiv; 2019.  \url{http://arxiv.org/abs/1909.01610}. Accessed August 31, 2023.

\item [32.] 	Lita LV, Rogati M, Lavie A. BLANC: learning evaluation metrics for MT. Proc Conf Hum Lang Technol Empir Methods Nat Lang Process - HLT 05. Vancouver, British Columbia, Canada: Association for Computational Linguistics; 2005. p. 740–747. doi: \url{http://doi.org/10.3115/1220575.1220668}.

\item [33.]  	Gao Y, Zhao W, Eger S. SUPERT: Towards New Frontiers in Unsupervised Evaluation Metrics for Multi-Document Summarization. Proc 58th Annu Meet Assoc Comput Linguist. Online: Association for Computational Linguistics; 2020. p. 1347–1354. doi: \url{http://doi.org/10.18653/v1/2020.acl-main.124}.

\item [34.] Grusky M, Naaman M, Artzi Y. Newsroom: A Dataset of 1.3 Million Summaries with Diverse Extractive Strategies. Proc 2018 Conf North Am Chapter Assoc Comput Linguist Hum Lang Technol Vol 1 Long Pap. New Orleans, Louisiana: Association for Computational Linguistics; 2018. p. 708–719. doi: \url{http://doi.org/10.18653/v1/N18-1065}.

\item [35.] Fabbri AR, Kryściński W, McCann B, Xiong C, Socher R, Radev D. SummEval: Re-evaluating Summarization Evaluation. Trans Assoc Comput Linguist. 2021;9:391–409. doi:  \url{http://doi.org/10.1162/tacl_a_00373}.

\item [36.]	Huemann Z, Lee C, Hu J, Cho SY, Bradshaw T. Domain-adapted large language models for classifying nuclear medicine reports. arXiv; 2023.  \url{http://arxiv.org/abs/2303.01258}. Accessed March 17, 2023.

\end{itemize}

\end{document}